\documentclass[journal]{IEEEtran}
\usepackage{xcolor,soul,framed} 
\usepackage{array}
\usepackage{mdwmath}
\usepackage{mdwtab}
\usepackage{booktabs}

\usepackage{times}
\usepackage{epsfig}
\usepackage{graphicx}
\usepackage{amsmath}
\usepackage{amssymb}
\usepackage{makecell,multirow,diagbox}
\usepackage[normalem]{ulem}
\usepackage{ulem}
\usepackage{comment}

\def\eg{\emph{e.g.}}
\def\ie{\emph{i.e.}}
\begin{document}
\bstctlcite{IEEEexample:BSTcontrol}
    \title{Object Discovery From a Single Unlabeled Image by Mining Frequent Itemset With Multi-scale Features}
    
  \author{Runsheng Zhang, 
      Yaping Huang, 
      Mengyang Pu, 
      Jian Zhang, Qingji Guan, Qi Zou, 
      Haibin Ling~\IEEEmembership{Member,~IEEE,} 
  \thanks{This work is supported by the Fundamental Research Funds for the Central Universities (2019YJS027, 2019JBZ104) and National Natural Science Foundation of China (61906013, 51827813). (Corresponding author:Yaping Huang.)}
  \thanks{R. Zhang, Y. Huang, Q. Zou, M. Pu, J. Zhang and Q. Guan are with the Beijing Key Laboratory of Traffic Data Analysis and Mining, Beijing Jiaotong University, Beijing 100044, China (e-mail:\{rszhang, yphuang, mengyangpu, jianzhang1, qzou\}@bjtu.edu.cn; qingjiguan@gmail.com).}
  \thanks{H. Ling is with the Department of Computer Sciences, Stony Brook University, Stony Brook, NY 11794  USA (e-mail: hling@cs.stonybrook.edu).}
  }


\markboth{ACCEPTED BY IEEE TIP
}{Roberg \MakeLowercase{\textit{et al.}}: High-Efficiency Diode and Transistor Rectifiers}

\maketitle

\begin{abstract}
The goal of our work is to discover dominant objects in a very general setting where only a single unlabeled image is given. This is far more challenge than typical co-localization or weakly-supervised localization tasks. To tackle this problem, we propose a simple but effective pattern mining-based method, called Object Location Mining (OLM), which exploits the advantages of data mining and feature representation of pre-trained convolutional neural networks (CNNs). Specifically, we first convert the feature maps from a pre-trained CNN model into a set of transactions, and then discovers frequent patterns from transaction database through pattern mining techniques. We observe that those discovered patterns, \ie, co-occurrence highlighted regions, typically hold appearance and spatial consistency. Motivated by this observation, we can easily discover and localize possible objects by merging relevant meaningful patterns. Extensive experiments on a variety of benchmarks demonstrate that OLM achieves competitive localization performance compared with the state-of-the-art methods. We also evaluate our approach compared with unsupervised saliency detection methods and achieves competitive results on seven benchmark datasets. Moreover, we conduct experiments on fine-grained classification to show that our proposed
method can locate the entire object and parts accurately, which
can benefit to improving the classification results significantly.
\end{abstract}


%
\IEEEpeerreviewmaketitle


\section{Introduction}

\IEEEPARstart{O}{bject} 
 discovery and localization is a fundamental computer vision problem, and it aims to discover and locate interesting objects within an image.
 Benefiting from the learning capability of deep convolutional neural networks and large-scale object bounding box annotations, object localization has achieved remarkable performance~\cite{Liu2016SSD,Redmon2016You}. However, it is expensive and labor-intensive to manually annotate bounding boxes on the large-scale dataset. This motivates the development of weakly supervised object localization methods, which only use image-level annotations to localize objects. Most existing weakly supervised object localization methods locate the objects by training a classification convolutional neural network with image-level annotations~\cite{Zhou2016Learning,Durand2017WILDCAT,XiaolinAdversarial2018}.
 \begin{figure}
\includegraphics[width=1.0\linewidth]{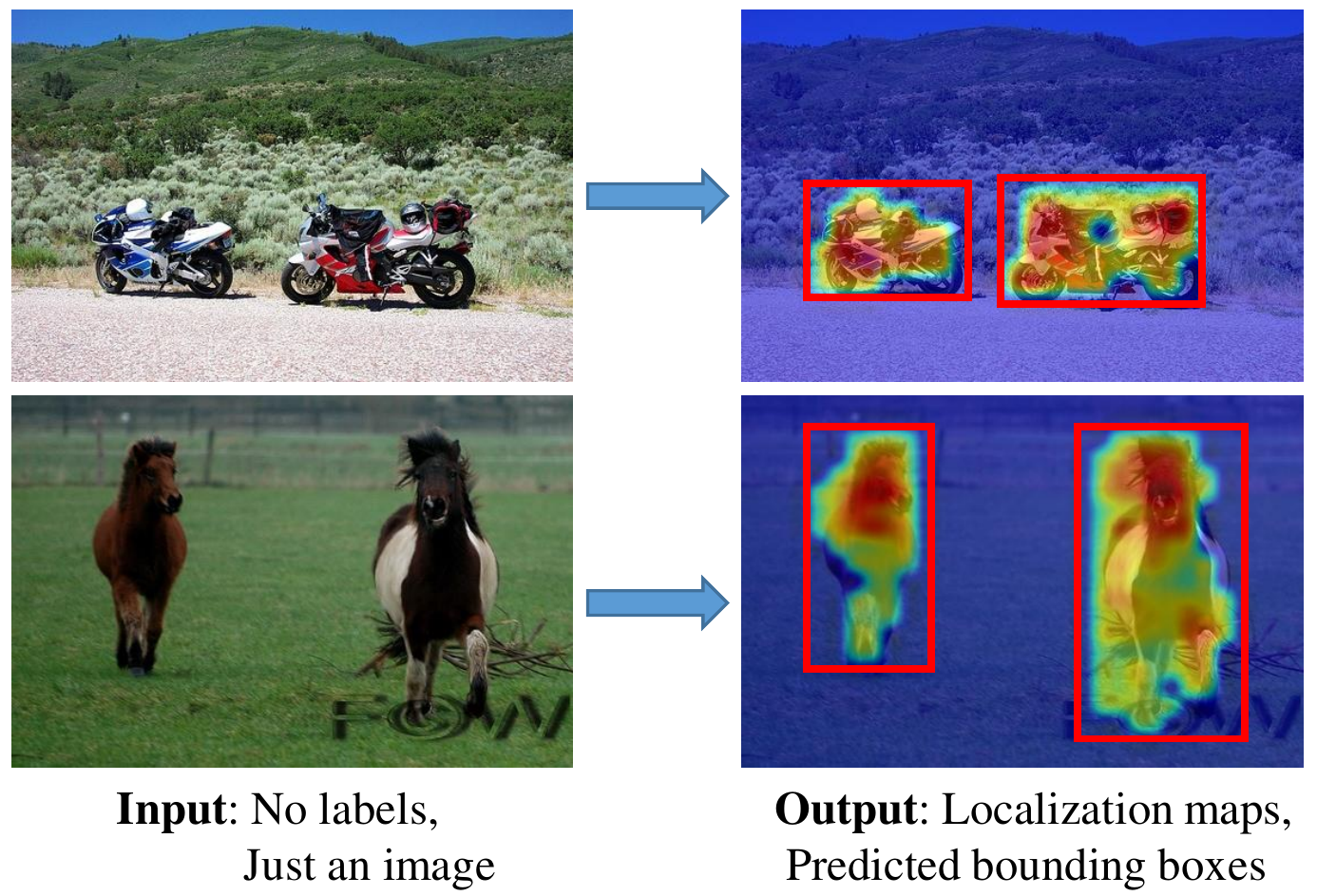}
\caption{Object localization from an unlabeled image. We tackle this problem in a challenging setting where only an image is given, without any type of annotations or even a set of images containing objects from the same category. Our proposed method discovers most dominant object instances (red bounding boxes) based on their localization maps. (Best viewed in color.)}
\label{fig:setting}
\end{figure}
 Recently, some works~\cite{Tang2014Co,Joulin2014Efficient,Cho2015Unsupervised,Li2016Image,Wei2017Deep} shift their attention to image co-localization problem, which assumes less supervision and requires a set of images containing objects from a common category. But most of these efforts usually need to first generate enormous object proposals. This may lead to high time consumption and the performance heavily depends on the quality of the proposals.

In this paper, we address object localization in a far more challenging and more realistic scenario where only a single unlabeled image is given. As illustrated in Figure~\ref{fig:setting}, the input of our method is just only a single image, without utilizing any annotations or even a single dominant class. In fact, this setting is reasonable due to the fact that most images in real life are usually with unknown categories or even without labels. Therefore, it is worth developing object localization methods from unlabeled images.

Recent works~\cite{Zeiler2014Visualizing,hariharan2015, Zhou2016Learning} have demonstrated that the pre-trained CNN models could provide powerful semantic representation for a given image.  
~\cite{Zhou2016Learning,Wei2017Selective} have also revealed that the convolutional activations usually fire at the same region that may be a general part of an object. Therefore, it is important and necessary to explore the potential of exploiting the pre-trained models: models trained for one task, but could be reuse in scenarios that different from their original purposes~\cite{zhou2016learnware}.~\cite{Wei2017Selective} first reuses the pre-trained model and employs a simple ``mean-threshold'' strategy to localize the objects. They add up the \emph{Pool-5} activation tensor through the depth direction and calculate the mean value as the threshold to decide which positions belong to objects. However, such simple strategy is only evaluated on fine-grained datasets. In fact, the key weakness of the localization maps generated by~\cite{Wei2017Selective} is that even if the activation responses of one region are higher than the mean value in only one single channel, the regions will also be fired, which leads to  the localization results are not very robust to the noisy background on complex images. Therefore, it is of great interest to develop more efficient approaches to automatically discover and utilize the location information of co-occurrence activations, which would be beneficial to identify integral object regions.

To tackle the above issues, this paper proposes a novel and simple approach, named Object Location Mining (OLM), to mine frequent patterns and discover objects from the pattern mining perspective. The success of our proposed method relies on two key foundations:
1) The CNN features extracted from the pre-trained model have powerful representation and provide abundant semantic and spatial information; 2) Pattern mining techniques can efficiently mine frequently-occurring activations, which often indicate the location of objects in one image.

Our proposed OLM first converts the deep features from the particular layer of a pre-trained CNN model into a set of transactions (\ie~a transaction database). Specifically, we propose an efficient transaction creation strategy: we extract the deep features from multiple convolutional layers and use a tunable threshold to select the descriptors that are used to convert to items. Benefiting from this strategy, we could retain the most useful information while discarding redundant information, which is critical to obtain a suitable input for a mining algorithm. Then we look for relevant but non-redundant patterns automatically through pattern mining techniques. Finally, we merge the selected patterns to generate a support map which represents the object regions. The experimental results show that the discovered regions are not only semantically consistent but also accurately cover the objects. Based on the mined objects, it allows for a more efficient part 
localization and can be easily integrated with fine-grained classification models.

Our proposed method is simple but effective, which does not need the training process. Thus, we do not need to design complex loss function and avoid collecting a large amount of annotations which is labor consuming. More importantly, compared with co-localization methods, we address object localization in more challenging scenarios where only a single unlabeled image is given. It is more reasonable in the practical scenario.

To the best of our knowledge, we propose the first usage of pattern mining for object localization. The main contributions are summarized as follows:

\begin{itemize}
  \item We propose an effective pattern mining-based method, named Object Location Mining (OLM), to localize object regions from a single unlabeled image, which 
  only utilizes pre-trained models and does not need any training or fine-tuning process. Our OLM demonstrates that incorporating pattern mining strategy with pre-trained models has potential to be a general object detector. 
   
  \item We propose an efficient transaction creation strategy to transform the location index of frequently-occurring convolutional activations into transactions, which is a key step to localize the possible object regions by using pattern mining techniques.
  
  \item Extensive experiments are conducted on a variety of datasets include four fine-grained datasets, Object Discovery dataset, ImageNet subsets and PASCAL VOC 2007 dataset. Our proposed method outperforms other unsupervised methods by a large margin. Moreover, we also evaluate the localization ability on unsupervised saliency detection and fine-grained classification task. Our method achieves competitive performance compared with the state-of-the-art methods.
  \end{itemize}

\section{Related Work}
In this section, we review related works
on object discovery, and its extension to fine-grained image classification and saliency detection, and the strategies of the pattern mining which
inspire us a novel solution to the proposed method.

\subsection{Object Localization From Unlabeled Data}
Object Localization from unlabeled data is challenging due to the fact that it does not depend on any auxiliary information rather than a given unlabeled image. Thus, many methods shift their attention to solve image co-localization problem~\cite{Faktor2012,Rubinstein2013Unsupervised,Joulin2014Efficient,Tang2014Co,Cho2015Unsupervised,Li2016Image,Wei2017Deep}.

Image co-localization requires a set of images containing objects from the same category. Some earlier image co-localization methods~\cite{Faktor2012,Rubinstein2013Unsupervised,Joulin2014Efficient,Cho2015Unsupervised} address this problem based on low-level features (\eg, SIFT, HOG).~\cite{Li2016Image} is the first to use the features from fully connected layer of a pre-trained CNN model to learn a common object detector. However, the spatial correlation of deep descriptors in convlutional layers are lost.~\cite{Zhou2015Object,Zhou2016Learning} demonstrate that the convolutional activations remain spatial and semantic information and have remarkable localization ability.~\cite{Wei2017Deep} utilizes the convolutional activations and further considers their correlations in an image collection to deal with image co-localization problem.

Since object localization from unlabeled data is a very challenging problem, there exist a limited number of comparable methods such as~\cite{Wei2017Selective}.~\cite{Wei2017Selective} extracts feature descriptors from the last max-pooling layer of a pre-trained VGG-16 model~\cite{Simonyan2014Very} and employs a simple ``mean-threshold'' strategy to locate the main objects in fine-grained images. Compared with~\cite{Wei2017Selective},  our pattern mining-based method considers a more reasonable and more general measurement only those areas that are frequently activated in many channels will be fired. Thus, our method is more robust to the noisy background and can mine more distinctive object regions.

\subsection{Pre-trained Models}
Current deep learning methods have achieved great success on a variety of visual recognition tasks~\cite{Liu2016SSD,Redmon2016You}. However, these deep frameworks still heavily rely on a large amount of training data and time-consuming training process. Therefore, some researches have recently turned to a new direction: model reuse~\cite{zhou2016learnware}. Obviously it would be appealing if a pre-trained model can be reused in a new domain that is different from its initial training purpose without training the model from scratch, and even without requiring to fine-tune the pre-trained model. Fortunately, benefiting from the large-scale dataset ImageNet, which consists of over 1.2 million images in over 1,000 categories, the pre-trained models have revealed powerful feature representation ability and can also be used as general object detectors~\cite{Zhou2016Learning}. Meanwhile, some recent works attempt to train deep models on larger image dataset collecting from web images~\cite{mahajan2018exploring}, which can further leverage the generalization of pre-trained models.

In this paper, our proposed OLM extracts the feature representations from pre-trained models without any training or fine-tuning process, which is similar with~\cite{Wei2017Selective,Wei2017Deep}. We further mine the frequently-occurring regions from activations of pre-trained models using pattern mining techniques. We demonstrate that OLM can be easily implemented and applied to object localization and saliency detection, and also transferred to other tasks, \eg, fine-grained classification.

\begin{figure*}[t]
\begin{center}
\includegraphics[width=0.92\linewidth]{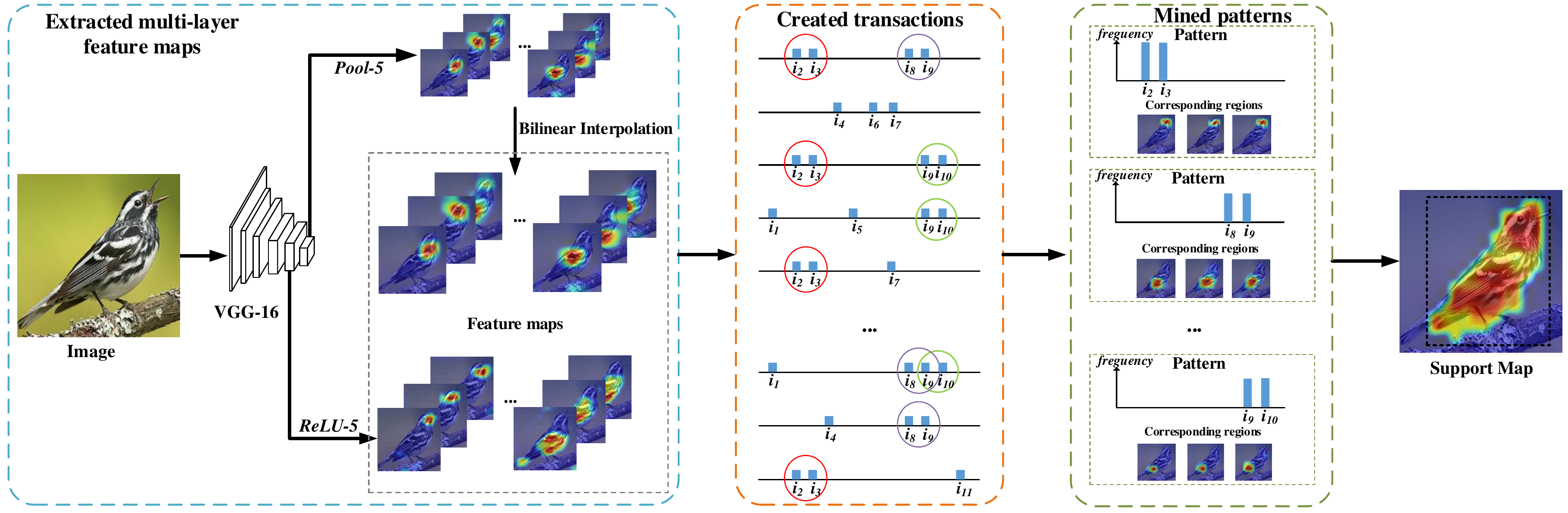}
\caption{Overview of our proposed OLM for object localization from an unlabeled image. First, feature maps (\emph{Pool-5} layer and \emph{ReLU-5} layer) are extracted from a VGG-16 model\cite{Simonyan2014Very} pre-trained on ImageNet, and feature maps from \emph{Pool-5} layer are resized to the same size of \emph{ReLU-5} layer by bilinear interpolation. Then, we transform those feature maps into a transaction database, thus meaningful patterns could be discovered from the transaction database through Object Mining. Finally, a support map is generated by merging the meaningful patterns.
}
\label{fig:overview}
\end{center}
\end{figure*}

\subsection{Fine-grained Image Classification}

Fine-grained image classification is one of the most fundamental and important task in computer vision, and a large amount of works have been developed in the past few years. Benefited from the advancement of deep learning, many works~\cite{krizhevsky2012imagenet,Simonyan2014Very,zhang2014part,xiao2015application,lin2015bilinear} learn more discriminative feature representation by leveraging deep CNNs, and achieve significant progress.

Since subtle visual differences mostly reside in local regions of parts, discriminative part localization is crucial for fine-grained image classification. There are numerous emerging works proceeding along part localization. \cite{zhang2014part,huang2016part,zhang2016spda,wei2018mask} learn accurate part localization models with manual object bounding boxes and part annotations. Considering that the annotations are laborious and expensive, some works~\cite{zhang2016weakly,zhang2016picking,he2017fine,xiao2015application,zhao2017diversified,liu2016fully,fu2017look,zheng2017learning} instead focus on how to exploit parts under a weakly-supervised setting with only image-level labels. \cite{zhang2016picking} proposes an automatic fine-grained classification method, incorporating deep convolutional filters with significant and consistent responses for both parts selection and representation.
Some of the above part localization-based methods~\cite{zhang2016weakly,xiao2015application,zhang2016picking,zhang2016detecting} usually require to firstly produce object or part candidates by selective search~\cite{uijlings2013selective}, which poses challenges to accurate part localization. 

Additionally, some weakly-supervised methods~\cite{sermanet2014attention,xiao2015application,zhao2017diversified,liu2016fully,fu2017look,zheng2017learning} use visual attention mechanism to automatically capture the informative regions. \cite{liu2016fully} employs a fully convolutional attention network to adaptively localize multiple parts simultaneously. Recent works~\cite{fu2017look, zheng2017learning} propose an end-to-end framework where part localization and feature learning mutually reinforce each other. Although promising results have been reported, it is highly difficult to train the models due to sophisticated alternative training procedures.

Since part localization is a key issue in fine-grained classification, it is natural to extend our Object Mining method to part localization. Therefore, we further propose an efficient part localization method. Our method does not need sophisticated training procedures. We can directly localize multiple fine-grained parts instead of selecting useful ones from enormous part proposals with high cost~\cite{simon2015neural}. 

\subsection{Saliency Detection}
Saliency detection aims to automatically discover and locate the most salient regions in images or videos that draw human attention. Depending on annotations used, saliency detection methods can be roughly divided into two categories: unsupervised methods and supervised methods. ~\cite{DBLP} and~\cite{In_Depth_Survey} provide the detailed reviews of saliency detection. Our work focuses on discovering the objects from an unlabeled image and thus is more related to unsupervised saliency detection methods, which mainly rely on different priors. 

One of the earliest works on unsupervised salient object detection is \cite{achanta2009frequency}, which calculates the color deviation from the average image color at the pixel level. Later MC~\cite{Jiang2013SaliencyMC} separates salient objects from the background via absorbing Markov chain.
\cite{yan2013hierarchical,margolin2013makes,cheng2013efficient,yang2013saliency} make use of high-level human perceptual knowledge, such as context, semantics and background. \cite{qin2015saliency} introduces a mechanism which is dependent on Cellular Automata to explore the salient object. \cite{Tong2015} attempts to detect the salient regions via bootstrap learning. WLRR~\cite{tang2016salient} and WMR~\cite{zhu2018saliency} use a ranking framework to discover salient objects. MDC~\cite{huang2017300} proposes minimum directional contrast as raw saliency metric. Recently, some researchers have shifted their attention to video saliency detection~\cite{tu2016real,xi2016salient,wang2017saliency}. STBP~\cite{xi2016salient} focuses on spatiotemporal background priors. MSTM~\cite{tu2016real} presents a real-time method based on the minimum spanning tree. \cite{wang2017saliency} solves the pixel-wise segmentation task as an energy minimization problem. 

The other branch of salency detection is supervised methods, which need pixel-level labels for training. For example, DRFI~\cite{Jiang2013SalientDRFI} integrates the regional contrast, regional property and regional backgroundness descriptors together to produce the saliency map. Recently, CNN models are gradually becoming the mainstream solution in saliency detection~\cite{wei2019f3net,fu2019deepside}.
For promoting the development of saliency community, \cite{fan2018SOC} explores a deeper insight into the SOD problem and proposes a new high-quality SOC dataset. At present, \cite{wang2017video,fan2019shifting} utilize deep neural networks to solve the video saliency detection.   
These approaches can achieve state-of-the-art performance due to encoding high-level semantic features.  

 However, the CNN-based methods strongly require large-scale annotations, which are not only labor-intensive but also computation expensive. Some methods explore a combination of deep learning and the unsupervised setting to solve the problem of saliency detection. \cite{Zhang2017Supervision,Zhang_2018_CVPR} use existing unsupervised salient models to generate pseudo ground-truth, and then train a CNN-based model to solve the problem. 

Compared to the aforementioned methods, our method does not need any training process, leading to a more practical solution. 

\subsection{Pattern Mining in Computer Vision}

Pattern mining techniques have been developed for several decades in the data mining community. Usually, a set of patterns is a combination of several elements and the distinctive information is captured. Inspired by this fact, many researchers investigate the problem of employing pattern mining to address computer vision tasks, including image classification~\cite{Li2014Mid,Fernando2014Mining}, image collection summarization~\cite{Rematas2015Dataset} and object retrieval~\cite{Fernando2014Mining}.

A key issue in pattern mining methods is how to transform an image into transaction data, which are suitable for pattern mining and maintain most of the corresponding discriminative information. 
Most early methods simply treat an individual visual word as an item in a transaction. Due to the sparsity of local bag-of-words (LBOW), LBOW is usually adopted as image representations in~\cite{QuackFLG07,YuanWY07}, then each visual word is treated as an item. However, this operation only considers the absence/presence of the visual word and may lead to information loss during transaction creation. To avoid the above issues,~\cite{Fernando2014Mining1} proposes a frequent local histogram method to represent an image with the histograms of patterns sets.
More recently,~\cite{Li2014Mid} is first to illustrate how pattern mining techniques are combined with the CNN features, a more appealing alternative than the hand-crafted features. In ~\cite{Li2014Mid}, a local patch is transformed into a transaction by treating each dimension index of a CNN activation as an item. 

Different from the previous works which need a set of same class images as input and focus on finding frequent visual words to generate better feature representations, the goal of our work is significantly different. We aim at finding the frequently-activated position information of feature maps to localize the possible object regions. Obviously, the previous works are more suitable for classification and recognition tasks. In contrast, we propose the first usage of pattern mining technique for object localization task.

\section{The Proposed Method}
In this section, we provide details of our OLM approach. The overview of the proposed method is illustrated in Figure~\ref{fig:overview}. First, we extract the feature maps from the \emph{Pool-5} and \emph{ReLU-5} layers of a 
VGG-16 model pre-trained on ImageNet. Then, the feature maps are converted into a set of transactions and the meaningful patterns are discovered by  
pattern mining techniques. Finally, we illustrate the details of how to merge the selected patterns to localize potential target regions.

\subsection{Notations and Terminology}
First, we introduce the data mining notations and terminology.
The concept of frequent itemset originally introduced by mining frequent itemsets from a transaction database for market data analysis~\cite{Agrawal1994Fast}. Formally,
let $I = \{{i_1},{i_2},...,{i_M}\}$ be a set of items. The database D consists of a set of $transactions$ in which each transaction $T$ is a subset of $I$,~\eg,  $D = \{ {T_1},{T_2},...,{T_N}\}$, and $T_1 = \{ {i_1},{i_2},{i_7}\}$. 
Given an itemset $P \subseteq I$, we calculate the frequency $K$ of the itemset $P$ in the transaction database $D$ by $K = \left| {\left\{ {T\left| {T \in D,P \subseteq T} \right.} \right\}} \right| $.
Then, we define the \emph{support} value of $P$ as:
\begin{equation}
supp(P) =\frac{K}{N} =\frac{{\left| {\left\{ {T\left| {T \in D,P \subseteq T} \right.} \right\}} \right|}}{N} \in \left[ {0,1} \right],
\end{equation}
where $\left| \cdot \right|$ measures the cardinality, and $N$ is the number of transactions in $D$. 
The itemset $P$ whose \textit{support} value is larger than a predefined threshold is considered as the \textit{frequent itemset}.
The itemset $P$ is the pattern we want to mine.

\begin{figure}[t]
\includegraphics[width=1\linewidth]{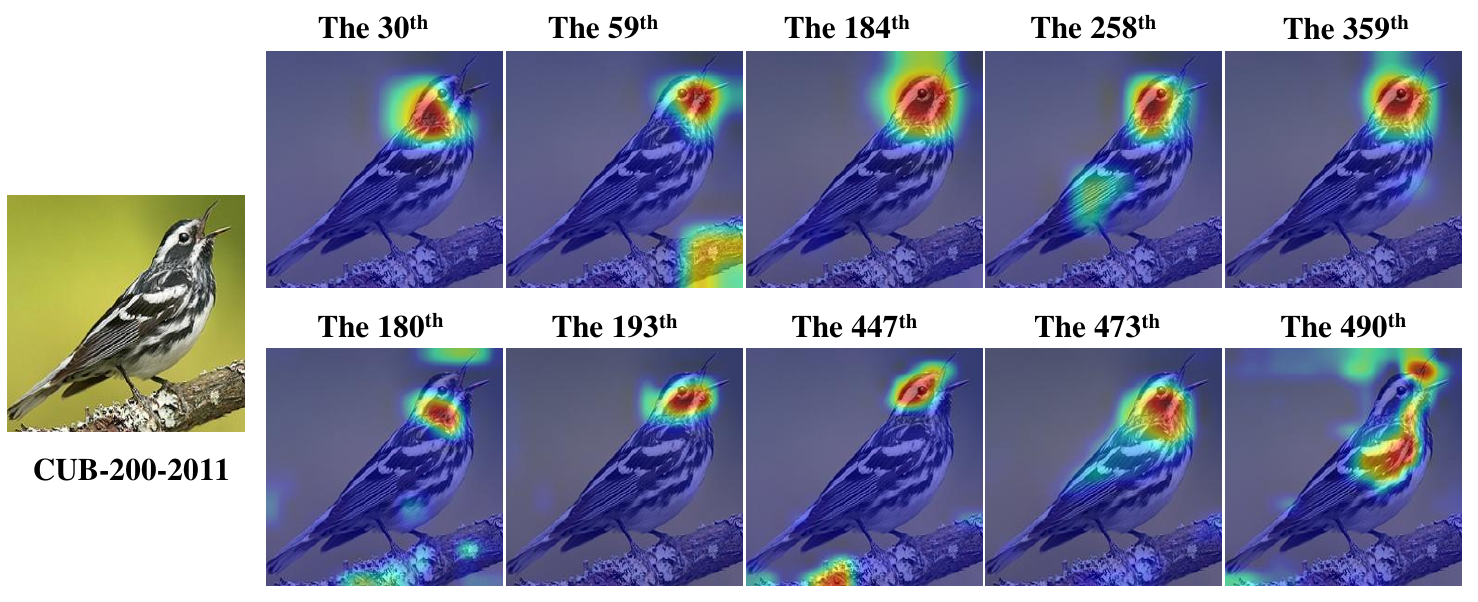}
\caption{Sampled feature maps of an image from CUB-200-2011 dataset. The first column is the input image, and the rest five feature maps of the first row are selected from \emph{Pool-5}, and the five feature maps of the second row are select from \emph{ReLU-5}. (Best viewed in color.)}
\label{fig:featuremap}
\end{figure}
\subsection{Extracting Multi-layer Feature Maps}
Recent works demonstrate that different convolutional layers learn different level features~\cite{Zeiler2014Visualizing}. Specifically, 
the deeper the convolutional layer, the better discriminative power it has. Thus, the general rule is that deeper convolutional layers are preferred to choose. In our work, the images are passed through a pre-trained VGG-16~\cite{Simonyan2014Very} model and the feature maps are obtained from \emph{Pool-5} and \emph{ReLU-5} layers. This strategy allows us to take original images with arbitrary sizes as input, and also alleviates the loss of useful information caused by only considering single layer activations.

\begin{figure*}
\begin{center}
\includegraphics[width=1\linewidth]{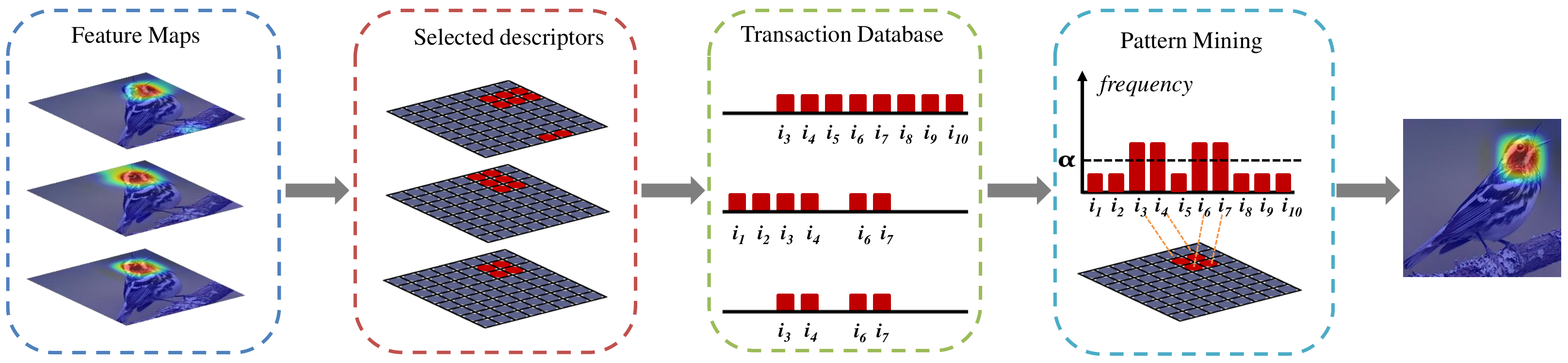}
\caption{The detailed process of pattern mining. 
We first select the descriptors and convert them to items which are utilized to transform the feature maps into a transaction database. 
$\left\{ {{i_1},{i_2},...,i_{10}}\right\}$ is the index set of the highlighted positions, \ie \@ the item set of the transactions. 
Then we statistically count the frequencies of each item and remain the items whose frequency are greater than $\alpha$. Finally, we mine the co-occurrence highlighted regions corresponding to the frequent patterns.
}
\label{fig:patternming}
\end{center}
\end{figure*}

We randomly select an image from the fine-grained dataset (CUB-200-2011~\cite{Wah2011The}) and visualize the local response regions of \emph{Pool-5} and \emph{ReLU-5} layers. As shown in Figure~\ref{fig:featuremap}, we can clearly observe that although the heatmaps of \emph{Pool-5} and \emph{ReLU-5} are not the same, most semantic parts of a bird are frequently activated at the same location in many feature channels.  
Moreover, these two layers can preserve more co-occurrence meaningful regions, which may be the corresponding parts of an object. However, some background regions are also activated. Fortunately, these regions only present in few channels and are distributed sparsely. That means not all activated areas in the feature maps are useful, so it is necessary to further mine the meaningful regions using pattern mining techniques. 

\subsection{From Feature Maps to Transactions}

It is the most critical step to convert data into a set of transactions while maintaining useful information  when applying pattern mining techniques to computer vision applications. 
In our method, each feature map is converted into a transaction denoted by $T$, and each position index activated from the feature map is considered as an item $i$. For example, if there are five positions fired in the $j\textrm{-}th$ feature map, the corresponding transaction $T_j$ would include five items, \ie $T_j = \left\{ {{i_1},{i_2},{i_3},{i_4},{i_5}}\right\}$. The index set of all activated positions from all feature maps, also known as an itemset, is denoted by $I = \left\{ {{i_1},{i_2},...,{i_m}}\right\}$. Each transaction is a subset of items $I$, \ie $T\subseteq I$. The set of all transactions (\ie all feature maps) is denoted by $D$. In our scenario, $N$ feature maps correspond to $N$ transactions, \ie $D = \left\{ {{T_1},{T_2},...,{T_N}}\right\}$.

In practice, an image with size $H\times W\times 3$~is fed into a pre-trained VGG-16 model~\cite{Simonyan2014Very}. Then we obtain 512 feature maps with size\@ $\frac{H}{32}\times \frac{W}{32}$ from \emph{Pool-5} and 512 feature maps with size $\frac{H}{16}\times \frac{W}{16}$ from \emph{ReLU-5} respectively. We resize \emph{Pool-5} feature maps to the same size with \emph{ReLU-5} by bilinear interpolation. Therefore, the size of each feature map is $\frac{H}{16}\times \frac{W}{16}$.

Next, we choose useful descriptors that can be converted into items. In~\cite{Li2014Mid}, a fixed threshold $k$ is used to select top $k$ activations from the full connected layer. However, this may result in the loss of some distinctive activations. In order to avoid this problem, we adopt a more appropriate and flexible strategy. More specifically, for each feature map, we calculate the mean value $\beta$ of the activation responses larger than $0$ as the tunable threshold. A position whose response magnitude is higher than $\beta$ is highlighted and the index will be converted into an item.

As noted in~\cite{Li2014Mid}, there are two strict requirements for applying pattern mining techniques: 1) only a small number of items can be included in one transaction; 2) only a set of integers can be recorded in one transaction. Fortunately, our proposed transaction conversion satisfies both. First, 
only small number of regions in each channel are activated. More importantly, the  threshold ensures us to select useful descriptors and discard the disturbing descriptors, in particular the activation responses of the background 
are low, and they will be abandoned. Therefore, it guarantees that the number of activations in each feature map is limited. Second, since we treat the indexes of the selected position in each feature map as items, this ensures that a transaction is recorded with a small set of integers. So our conversion strategy helps us to further mine possible objects successfully using pattern mining techniques.

\subsection{Mining Patterns}
Given a set of transaction database $D$, we apply the Apriori algorithm~\cite{Agrawal1994Fast} to find the frequent items. For a given minimum support threshold $\alpha$, an itemset $P$ is considered as frequent if $supp(p)\geqslant \alpha$. In other words, the support threshold determines which patterns will be mined.

The overall process of pattern mining is shown in Figure~\ref{fig:patternming}. First, we convert feature maps to a transaction database using the tunable threshold strategy. 
Note that the selected positions and the discarded positions are represented by red and blue dots, respectively. Then, we calculate the frequency and keep the items whose frequency is larger than $\alpha$. For example, we can observe that the frequency of $\left\{ {{i_3},{i_4},{i_6},{i_7}}\right\}$ is greater than $\alpha$, so we denote these items as a frequent pattern $P= \left\{ {{i_3},{i_4},{i_6},{i_7}}\right\}$. Thus, all frequent patterns are mined and the corresponding regions are discovered. 

\subsection{Selecting and Merging Patterns}

Although patterns are mined in the previous section, some of them may be isolated. Therefore we need to select the optimal patterns for object localization. Here we introduce our selection strategy: \emph{spatial continuity}. In our method, a mined pattern corresponds to a region in one image. Since one target object is spatially continuous in one image, the object regions represented by selected patterns should also be spatially continuous. For example, if we have mined two patterns $T_1 = \{ {i_1},{i_2},{i_7}\}$ and $T_2 = \{ {i_3},{i_4},{i_9}\}$, then these two patterns can be connected and merged into one region. It is because that $T_1$ and $T_2$ are two index sets of the activated positions in our method, and the positions represented by $i_2$ and $i_3$ are spatially adjacent. In addition, we discover that the isolated regions usually belong to the background of an image. Thus, we attempt to select the largest connected component based on the mined patterns to cover the entire target object as much as possible.

Subsequently, we merge the selected patterns to generate a support map $S$ for each image. Specifically, $S$ is defined as $s(x,y)$, where $s(x, y)$ is the frequency of an item represented by its position $(x, y)$. Note that the support map $S$ is generated by relevant and non-redundant patterns. The size of support map is same with the feature map. To obtain the support map with the same size as the original image, we upsample the support map by bilinear interpolation. 
Finally, we take the bounding box that covers the largest connected component in the support map, which is similar to CAM~\cite{Zhou2016Learning}. More significantly, the higher value $s(x, y)$ of the position, the more likely its corresponding region contains a part of a target object.

\section{Object Localization Experiments}

In this section, we evaluate our proposed OLM on object localization task. 

\subsection{Implementation Details and Datasets}
\textbf{Implementation details}. In our experiments, we extract feature maps from the convolutional layers (\emph{ReLU-5} and \emph{Pool-5}) of the publicly available pre-trained VGG-16 model~\cite{Simonyan2014Very}. Object Mining is implemented with MatConvNet~\cite{Vedaldi2015MatConvNet}.

\textbf{Datasets}: In order to evaluate the performance of the proposed approach, we conduct a set of qualitative and quantitative experiments on a variety of benchmark datasets. Specifically
CUB-200-2011~\cite{Wah2011The}, Stanford Dog~\cite{Khosla_l}, FGVC-Aircrafts~\cite{Maji2013Fine}, Standford Cars-196~\cite{Krause20133D}, Object Discovery dataset~\cite{Rubinstein2013Unsupervised}, ImageNet Subsets dataset\cite{Li2016Image} and PASCAL VOC 2007~\cite{pascal-voc-2007} are used for object localization. 

\begin{itemize}
\item Fine-grained datasets: CUB-200-2011 (200 classes, 11,788 images)~\cite{Wah2011The}, Stanford Dogs (120 classes, 20,580 images)~\cite{Khosla_l}, Stanford Cars-196 (196 classes, 16,185 images)~\cite{Krause20133D} and FGVC-Aircrafts (100 classes, 10,000 images)~\cite{Maji2013Fine}.

\item The Object Discovery Dataset. The Object Discovery
dataset~\cite{Rubinstein2013Unsupervised} contains 15k images in three categories: airplane, car, and horse. To compare with previous co-localization method fairly, we use a subset of the dataset containing 100 images for each category. There are 18, 11, 7 outliers for three categories, respectively.

\item The ImageNet Subsets: We follow other co-localization method to chose six subsets of the ImageNet dataset~\cite{Imagenet2009}. These subsets are held-out categories from the 1000-label ILSVRC classification. That is to say, these subsets are “unseen” by pretrained CNN models.
  
\item PASCAL VOC 2007 Dataset: The PASCAL VOC
2007 dataset~\cite{pascal-voc-2007} is one of the most popular datasets for computer vision tasks such as classification, detection, and segmentation, and it contains 5,011 images in 20 object categories. Following the previous methods~\cite{Joulin2014Efficient,Cho2015Unsupervised}, we evaluate our method on two sizes of sets: PASCAL07-6x2 and PASCAL07-all. The PASCAL07-6x2 subset consists of all images from 6 classes (aeroplane, bicycle, boat, bus, horse, and motorbike) of train+val dataset from the left and right aspect. For PASCAL07-all, we use all images in the train+val set (discarding images that only contain object instances marked as difficult or truncated).
\end{itemize} 

\textbf{Evaluation metrics}: Following previous image localization works~\cite{Cho2015Unsupervised,Wei2017Deep},~\cite{Wang_2015_ICCV}, we use the correct localization (CorLoc) to evaluate the proposed method. 
According the PASCAL-criterion, the CorLoc is defined as: $\frac{area(B_{p}\cap B_{gt})}{area(B_{p}\cup B_{gt})} > 0.5$, where $B_{p}$ is the predicted box and $B_{gt}$ is the ground-truth box.

\subsection{Fine-Grained datasets}

We conduct experiments on fine-grained datasets following some recently weakly-supervised methods \cite{Zhou2016Learning,XiaolinAdversarial2018,Zhang2018ECCV} and unsupervised method~\cite{Wei2017Selective}. Table~\ref{tab:tab1} shows the comparison with the state-of-the-art on four fine-grained datasets. The CorLoc arrives at 80.45\%, 80.70\%, 94.94\% and 92.51\% on CUB-200-2011, Stanford Dog, FGVC-Aircrafts and Standford Cars-196, respectively.

\begin{table}[htbp]
\footnotesize
\centering
\renewcommand\arraystretch{1.2}
\setlength{\tabcolsep}{10pt}
\begin{tabular}{|c|c|c|c|}
\hline
Dataset	& Method & Supervision & FullSet\\ 
\hline 
\hline
\multirow{7}*{CUB-200-2011}& \cite{Zhou2016Learning}& weakly & 59.00* \\
\cline{2-4}
 & \cite{Choe_2019_CVPR} & weakly &62.29* \\
\cline{2-4}
 & \cite{XiaolinAdversarial2018} & weakly &54.08* \\
 \cline{2-4}
 & \cite{Zhang2018ECCV} & weakly &56.33* \\
\cline{2-4}
& \cite{Xiang2017}& weakly& 65.52\\
\cline{2-4}
&\cite{Xu2016Friend}& weakly & 56.51 \\
\cline{2-4}
&\cite{Cho2015Unsupervised} & co-localization & 69.37 \\
\cline{2-4}
&\cite{Wei2017Selective}& w/o &  76.79 \\  
\cline{2-4}
& Ours  &w/o &  \textbf{80.45} \\  
\hline\hline
\multirow{4}*{Stanford Dogs} & \cite{Xu2016Friend}  & weakly &  66.68 \\
\cline{2-4}
&\cite{Cho2015Unsupervised} & co-localization &36.23 \\
\cline{2-4}
& \cite{Wei2017Selective} & w/o &78.86 \\
\cline{2-4}
 & Ours & w/o  &\textbf{80.70} \\
\hline
\hline 
\multirow{3}*{FGVC-Aircrafts}& \cite{Cho2015Unsupervised}& co-localization &42.91 \\
\cline{2-4}
& \cite{Wei2017Selective}  &w/o&94.91 \\
\cline{2-4}
& Ours & w/o  &\textbf{94.94} \\
\hline\hline
\multirow{3}*{Stanford Cars-196}& \cite{Cho2015Unsupervised} & co-localization &\textbf{93.05} \\
\cline{2-4}
& \cite{Wei2017Selective}  &w/o&90.96 \\
\cline{2-4}
& Ours &w/o&92.51 \\
\hline
\end{tabular}
\caption{Comparisons with state-of-the-art methods on four fine-grained datasets. Note that * denotes the test set of the full CUB-200-2011 dataset.}
\label{tab:tab1}
\end{table}

Compared with our method, only SCDA~\cite{Wei2017Selective} is proposed in the same setting. As shown in Table~\ref{tab:tab1}, we can observe that our method significantly exceeds SCDA on all four datasets (80.45\% \textit{vs} 76.79\% on CUB-200-2011, 80.70\% \textit{vs} 78.86\% on Dogs, 94.94\% \textit{vs} 94.91\% on Aircrafts, 92.51\% \textit{vs} 90.96\% on Cars-196). Moreover, our method is only slightly lower than co-localization method \cite{Cho2015Unsupervised} (92.51\% \textit{vs} 93.05\%) on Cars-196.

Besides, Table~\ref{tab:tab1} also shows that OLM outperforms some recent weakly supervised methods by a large margin on four fine-grained datasets. This result may be due to the following reason. Since only image labels can be used in weakly supervised methods, most of them are trained based on a classification neural network. Therefore, the generated discriminative areas are only suitable for classification, but may not be optimal for localization. \ie~the located areas are small or sparse regions instead of the whole object regions. In contrast, we directly reuse the pre-trained model and do not fine-tune on a specific dataset, which is beneficial for localizing object regions when incorporating powerful pattern mining techniques. We believe that our work can bring new insight into solving the localization problem.

\subsection{The Object Discovery dataset}

We further evaluate the performance of our proposed OLM on the Object Discovery dataset. The results are presented in Table~\ref{tab:object discovery}. Note that~\cite{Rubinstein2013Unsupervised,Tang2014Co,Wei2017Deep,Cho2015Unsupervised} are co-localization methods, which utilize a set of images containing the  objects from the same category.

In Table~\ref{tab:object discovery}, we can observe that OLM approach outperforms unsupervised localization method~\cite{Wei2017Selective} by 2.6\% (85.80\% \textit{vs} 83.20\%). Compared with co-localization methods, we also obtain a significant improvement, \ie~ 1.61\%~\cite{Cho2015Unsupervised}, 9.22\%~\cite{Tang2014Co} and 10.64\%~\cite{Rubinstein2013Unsupervised}. Our method is only lower than~\cite{Wei2017Deep}, which needs a set of images while only one single image is needed in our experiments. In particular, on the ``horse'' category, which is the most challenging subcategory due to multi-targets and complex background, our method achieves the new state-of-the-art compared with all other methods. These results suggest that our method is reasonably robust to the complex scenarios. Some examples are illustrated in Figure~\ref{fig:visualize}(a). 
\begin{table}[ht]
\footnotesize
\renewcommand\arraystretch{1.2}
\setlength{\tabcolsep}{7.5pt}
\begin{tabular}{|c|c|c|c|c|c|} 
\hline 
Methods&Supervision&Airplane&Car&Horse&\textbf{Mean}\\
\hline 
\hline 
\cite{Rubinstein2013Unsupervised}& co-localization  &74.39&87.64&63.44&75.16\\
\cite{Tang2014Co}& co-localization &71.95&93.26&64.52&76.58\\
\cite{Cho2015Unsupervised}& co-localization&82.93&94.38&75.27&84.19\\
\cite{Wei2017Deep}& co-localization&91.46&95.51&77.42&88.13\\
\hline 
\hline
\cite{Wei2017Selective} & w/o&87.80&86.52&75.37&83.20\\
Ours&w/o&89.02&89.88&78.49&85.80\\
\hline 
\end{tabular}
\caption{Comparisons of CorLoc (\%) on Object Discovery.} 
\label{tab:object discovery}
\end{table}

\begin{table}[ht]
\footnotesize
\renewcommand\arraystretch{1.2}
\setlength{\tabcolsep}{3.4pt}
\begin{tabular}{|c|c|c|c|c|c|c|c|} 
\hline 
Methods&Chipmunk&Rhino&Stoat&Racoon&Rake&Wheelchair&Mean\\
\hline  
\hline 
\cite{Cho2015Unsupervised}&26.6&81.8&44.2&30.1&8.3&35.3&37.7\\
\cite{Li2016Image} &44.9&81.8&67.3&41.8&14.5&39.3&48.3\\
\cite{Wei2017Deep}&70.3&93.2&80.8&71.8&30.3&68.2&69.1\\
\hline 
\hline
\cite{Wei2017Selective} &32.3&71.6&52.9&34.0&7.6&28.3&37.8\\
Ours &67.4&81.7&63.3&56.5&44.9&46.9&60.1\\
\hline 
\end{tabular}
\caption{Comparisons of CorLoc (\%) on ImageNet unseen subsets. Note that~\cite{Cho2015Unsupervised,Li2016Image,Wei2017Deep} are co-localization methods. Only ~\cite{Wei2017Selective} and our method localize the objects from an unlabeled image.}
\label{tab:imagenet}
\end{table}

\subsection{Unseen classes apart from ImageNet}
To demonstrate the generalization capabilities of our proposed OLM approach, 
we follow co-localization methods \cite{Cho2015Unsupervised,Li2016Image,Wei2017Deep} to conduct experiments on six subsets of the ImageNet. Note that, these subsets are “unseen” by pretrained CNN models. Table~\ref{tab:imagenet} presents the CorLoc metric on these subsets.

In Table~\ref{tab:imagenet}, we can see that OLM outperforms~\cite{Wei2017Selective} by a large margin (60.1\% vs 37.8\%) under the same setting. Compared with co-localization methods, our proposed OLM approach still significantly improves over two co-localization methods by about 22.4\%~\cite{Cho2015Unsupervised} and 11.8\%~\cite{Li2016Image} respectively. OLM is only lower than the recent co-localization method~\cite{Wei2017Deep}, but we achieve the best accuracy of 44.9\% and outperform~\cite{Wei2017Deep} by 14.6\% on the most difficult category (\ie, rake). The competitive results on the challenging ImageNet subsets demonstrate that incorporating pattern mining techniques can efficiently mine target objects from unlabeled data in real-world applications. We visualize some localization results of our OLM approach on the six subsets in Figure~\ref{fig:visualize}(b). It can be seen that our approach obtains decent object localization results in many challenging scenarios.

\begin{table*}[ht]
\footnotesize
\centering
\renewcommand\arraystretch{1.2}
\setlength{\tabcolsep}{3.6pt}

\begin{tabular}{|c|c|c|c|c|c|c|c|c|c|c|c|c|c|c|c|c|c|c|c|c|c|}
\hline 
Methods&aero&bike&bird&boat&bottle&bus&car&cat&chair&cow&table&dog&horse&mbike&person&plant&sheep&sofa&train&tv&Mean\\
\hline  
\hline 
\cite{Joulin2014Efficient}&32.8&17.3&20.9&18.2&4.5&26.9&32.7&41.0&5.8&29.1&34.5  &31.6&26.1&40.4&17.9&11.8&25.0&27.5&35.6&12.1&24.6\\
\cite{Cho2015Unsupervised}&40.4&32.8&28.8&22.7&2.8&48.4&58.7&41.0&9.8&32.0&10.2&41.9&51.9&43.3&13.0&10.6&32.4&30.2&52.7&21.8&31.3\\
\cite{Li2016Image}&73.1&45.0&43.4&22.7&6.8&53.3&58.3&45.0&6.2&48.0&14.3&47.3&69.4&66.8&24.3&12.8&51.5&25.5&65.2&16.8&40.0\\
\cite{wang2018simultaneously}&42.2&31.7&36.7&20.0&16.2&36.2&40.2&51.9&15.1&40.0&12.6&44.6&31.5&38.9&31.3&19.7&54.8&28.7&43.9&8.9&32.2\\
\cite{Wei2017Deep}&67.3&63.3&61.3&22.7&8.5&64.8&57.0&80.5&9.4&49.0&22.5&72.6&73.8&69.0&7.2&15.0&35.3&54.7&75.0&29.4&46.9\\
\hline 
\hline
\cite{Wei2017Selective}&54.4  &27.2&43.4&13.5 &2.8&39.3&44.5 &48.0&6.2&32.0&16.3  &49.8&51.5&49.7&7.7&6.1&22.1&22.6&46.4&6.1&29.5\\
Ours&66.7&36.4&50.4&30.0&11.8&53.5&49.4&44.3&18.9&38.9&27.7&53.1&53.0&62.0&12.9&8.2&28.8&30.0&69.9&12.6&37.9\\
\hline 
\end{tabular}
\caption{Comparisons of CorLoc (\%) on PASCAL07-all.  Note that~\cite{Joulin2014Efficient,Cho2015Unsupervised,Li2016Image,Wei2017Deep,wang2018simultaneously} are co-localization methods. Only ~\cite{Wei2017Selective} and our method localize the objects from a single unlabeled image.}
\label{tab:voc2007-all}
\end{table*}

\begin{table*}[ht]
\footnotesize
\centering
\renewcommand\arraystretch{1.2}
\setlength{\tabcolsep}{5.6pt}
\begin{tabular}{|c|cc|cc|cc|cc|cc|cc|c|} 

\hline 
 \multirow{2}{*}{Method}&\multicolumn{2}{c|}{aeroplane}&\multicolumn{2}{c|}{bike}&\multicolumn{2}{c|}{boat}&\multicolumn{2}{c|}{bus}&\multicolumn{2}{c|}{horse}&\multicolumn{2}{c|}{mbike}&\multirow{2}{*}{Mean}\\
&L&R&L&R&L&R&L&R&L&R&L&R&\\
\hline  
\hline 
\cite{Tang2014Co}&41.86&51.28&25.00&24.00&11.36&11.63&38.10&56.52&43.75&52.17&51.28&64.71&39.31\\
\cite{Cho2015Unsupervised}&62.79&66.67&54.17&56.00&18.18&18.60&42.86&69.57&70.83&71.74&69.23&44.12&53.73\\

\cite{niu2017knowledge}&64.28&87.17&35.41&39.13&16.66&30.00&42.86&60.86&46.80&53.33&59.45&75.75&50.97\\
\cite{wang2018simultaneously}&62.79&71.79&41.67&34.00&24.91&39.53&47.62&81.96&64.58&71.74&69.23&64.41&56.19\\
\hline 
\hline
\cite{Wei2017Selective}&67.44&74.36&33.30&38.00&11.36&20.93&33.33 &52.17&62.50&67.39&48.72 &76.47&48.83\\
Ours&76.74&84.62&43.75&46.00&25.00&32.56&52.38&60.87&66.67&67.39&64.11&79.41& 58.29\\
\hline 
\end{tabular}
\label{tab:voc2007-cx2}
\caption{Comparisons of CorLoc (\%) on PASCAL07-6x2. Note that~\cite{Tang2014Co,Cho2015Unsupervised,niu2017knowledge,wang2018simultaneously} are co-localization methods. Only ~\cite{Wei2017Selective} and our method localize the objects from a single unlabeled image.}
\label{tab:voc2007-6x2}
\end{table*}

\subsection{PASCAL VOC 2007 dataset}
We further evaluate our method on the PASCAL VOC 2007. We observe that many images consists of multiple object instances on PASCAL VOC 2007 dataset. But the generation of bounding box described in Section III-E focuses on single-object images. More importantly, our OLM does not require any type of annotations, thus we adopt a quite understandable way to generate the bounding boxes. We first select a pixel that is highlighted in the support map, and then we can obtain a bounding box that covers all the connected pixels of this connected component. If there are other pixels that are also highlighted but not be covered, we repeat the same operations to generate more bounding boxes until all the pixels are processed. Therefore, all the connected components are reserved, and each component corresponds to an object bounding box.

Since co-localization or weakly object localization methods only predict $C$ boxes in one image, where $C$ is the number of ground-truth object classes, we also need to reserve the same number of predicted boxes to give a fair comparison. However, the class labels of the images are unknown in our task, so after getting predicted boxes, we sort the boxes in descending order according to the pixel number of each box and remain $C$ boxes in one image for evaluation. 

\subsubsection{PASCAL07-all}
 Table~\ref{tab:voc2007-all} reports the results on PASCAL07-all. From the results, we can see that our proposed method outperforms unsupervised localization method~\cite{Wei2017Selective} by 8.4\% (37.9\% vs 29.5\%). Compared with co-localization method~\cite{Joulin2014Efficient,Cho2015Unsupervised,Li2016Image,Wei2017Deep}, our OLM approach also achieve competitive performance. However, for the challenging PASCAL VOC  dataset, our proposed OLM is still far behind the weakly object localization methods, e.g., Proposal Cluster Learning (PCL) method \cite{tang2018pcl} currently achieves a promising CorLoc of 66.6\% on trainval set.
 
 \begin{table}[t]
\footnotesize
\centering
\renewcommand\arraystretch{1.2}
\setlength{\tabcolsep}{8pt}
\begin{tabular}{|c|c|c|c|} 
\hline 
Dataset&\#GT boxes&\#predicted boxes&CorLoc\\
\hline 
\multirow{5}*{PASCAL07-all}&\multirow{5}*{3$ \sim $4 }&All (6$ \sim $7)&40.94\\
\cline{3-4}
&&3&40.20\\
\cline{3-4}
&&2&38.82\\
\cline{3-4}
&&C (1 $\sim$ 2) &37.90\\
\cline{3-4}
&&1&35.46\\
\cline{3-4}
\hline
\multirow{3}*{PASCAL07-6X2} &\multirow{3}*{1$ \sim $2}&All (4$ \sim $5)&60.26\\
\cline{3-4}
&&2&58.83\\
\cline{3-4}
&&C (1 $\sim$ 2)&58.29\\
\cline{3-4}
&&1&57.36\\
\hline
\end{tabular}
\caption{The CorLoc under the different predicted boxes on PASCAL VOC 2007. } 
\label{tab:Statistics}
\end{table}
 
\subsubsection{PASCAL07-6x2}
Table \ref{tab:voc2007-6x2} shows the evaluation results on the PASCAL07-6x2 dataset. We can see that even without any labels, our proposed method achieves promising results, which outperforms all other state-of-the-art co-localization methods~\cite{Tang2014Co,Cho2015Unsupervised,niu2017knowledge,wang2018simultaneously}. In addition, compared to~\cite{Wei2017Selective}, our approach can achieve 9.46\% performance gain. It also demonstrates the overall effectiveness of our approach.
\subsubsection{Effects of the number of predicted boxes} Since our proposed method can generate more than one predicted box in an image, we summarize the statistics of our predicted boxes and ground truth boxes on both PASCAL07-6x2 and PASCAL07-all datasets. PASCAL07-all contains about 3-4 instances per image on average, while most images in PASCAL07-6x2 contain 1-2 instances. As shown in Table \ref{tab:Statistics}, we can see that the number of our predicted boxes is not much different with that of ground truth boxes.

\begin{figure*}[ht]
\centering
\includegraphics[width=0.9\linewidth]{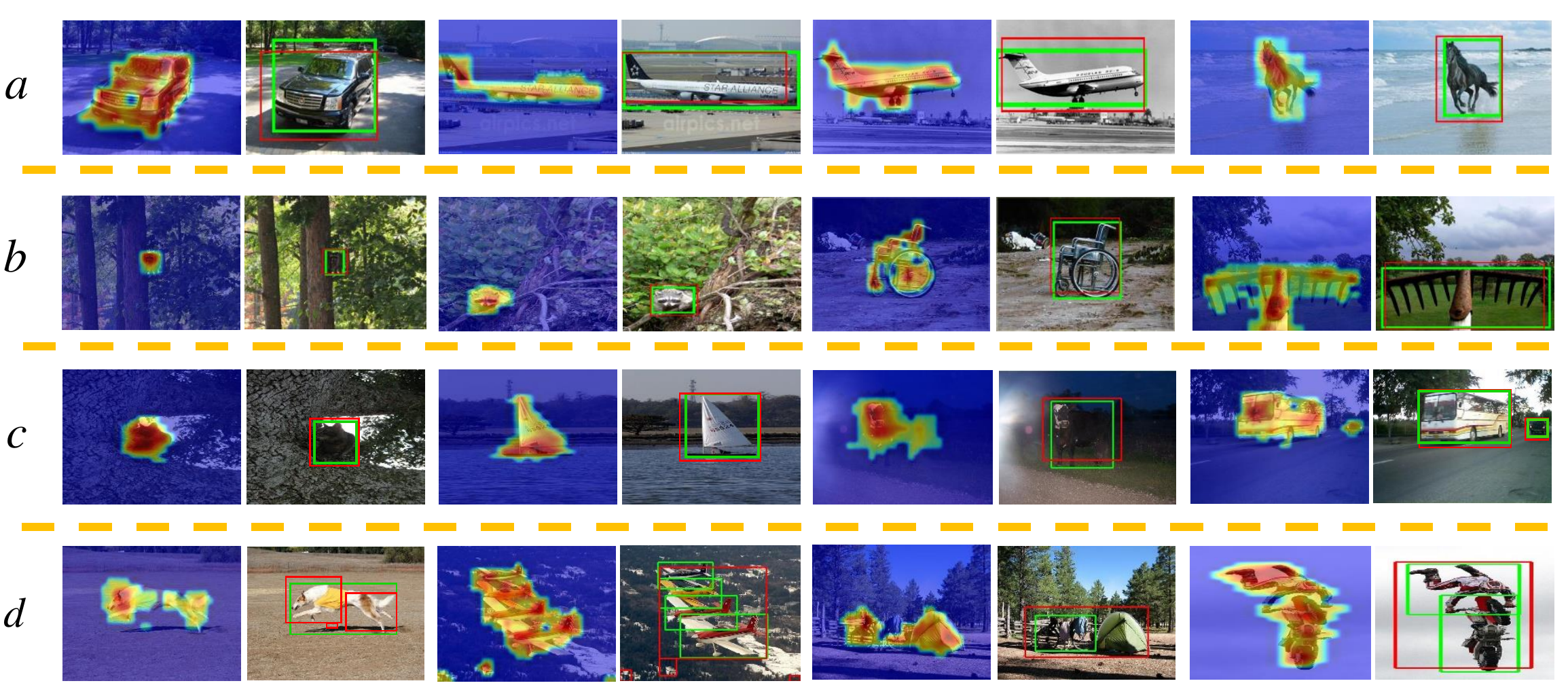}
\caption{Random samples of the predicted object localization bounding box and corresponding support map. (a) The Object Discovery dataset. (b) The ImageNet subsets dataset. (c) The PASCAL VOC 2007 data set. (d) Some failed localizations on the PASCAL VOC 2007 trainval set. The ground-truth bounding boxes are marked as the green rectangle, while the predicted ones are marked in the red rectangle. (Best viewed in color.)}
\label{fig:visualize}
\end{figure*}

In addition, we evaluate our method in terms of the CorLoc under the different predicted boxes to make our evaluation more comprehensively. ``The predicted boxes'' means how many generated boxes are remained for evaluation. We sort the boxes in descending order according to the pixel number of each box. ``All''means that all generated boxes are reserved. ``C'' means that only $C$ boxes are remained in one image, where $C$ is the number of ground-truth object classes. ``3'', ``2'' , and ``1'' refer to the number of the remaining boxes. The result is given in Table \ref{tab:Statistics}.
From the results, we can see that the CorLoc is relatively stable for the different numbers of boxes and consistently outperforms SCDA and most co-localization methods. Generally, our proposed method achieves a comparable performance only using a few predicted boxes.

\subsubsection{Discussion}
Furthermore, in Figure~\ref{fig:visualize}(c), we visualize some localization results of our OLM approach on PASCAL VOC 2007. It can be seen that our approach obtains decent object localization results in many challenging scenarios. Specifically, it discovers the precise objects even they are very similar to the background or in a complex background, \eg, the black cat on the tree. More surprisingly, although some images contain multiple objects, OLM can still effectively localize the target objects separately. However, some failure results are also shown in Figure~\ref{fig:visualize}(d). We can observe that the possible object regions have been highlighted successfully, but under the CorLoc metric, these will be judged as failure cases. This is mainly caused by errors during the generation of bounding boxes. Some better methods to generate bounding boxes from the support map can be considered to further improve the performance in the future work.

\subsection{Further Analysis}
This section evaluates the effects of the parameter $\alpha$ and the strategy of convolutional feature combination.

\subsubsection{The necessity of pattern mining strategy} 
\begin{figure}[t]
\setlength{\abovecaptionskip}{0.15cm}
\centering
\includegraphics[width=0.9\linewidth]{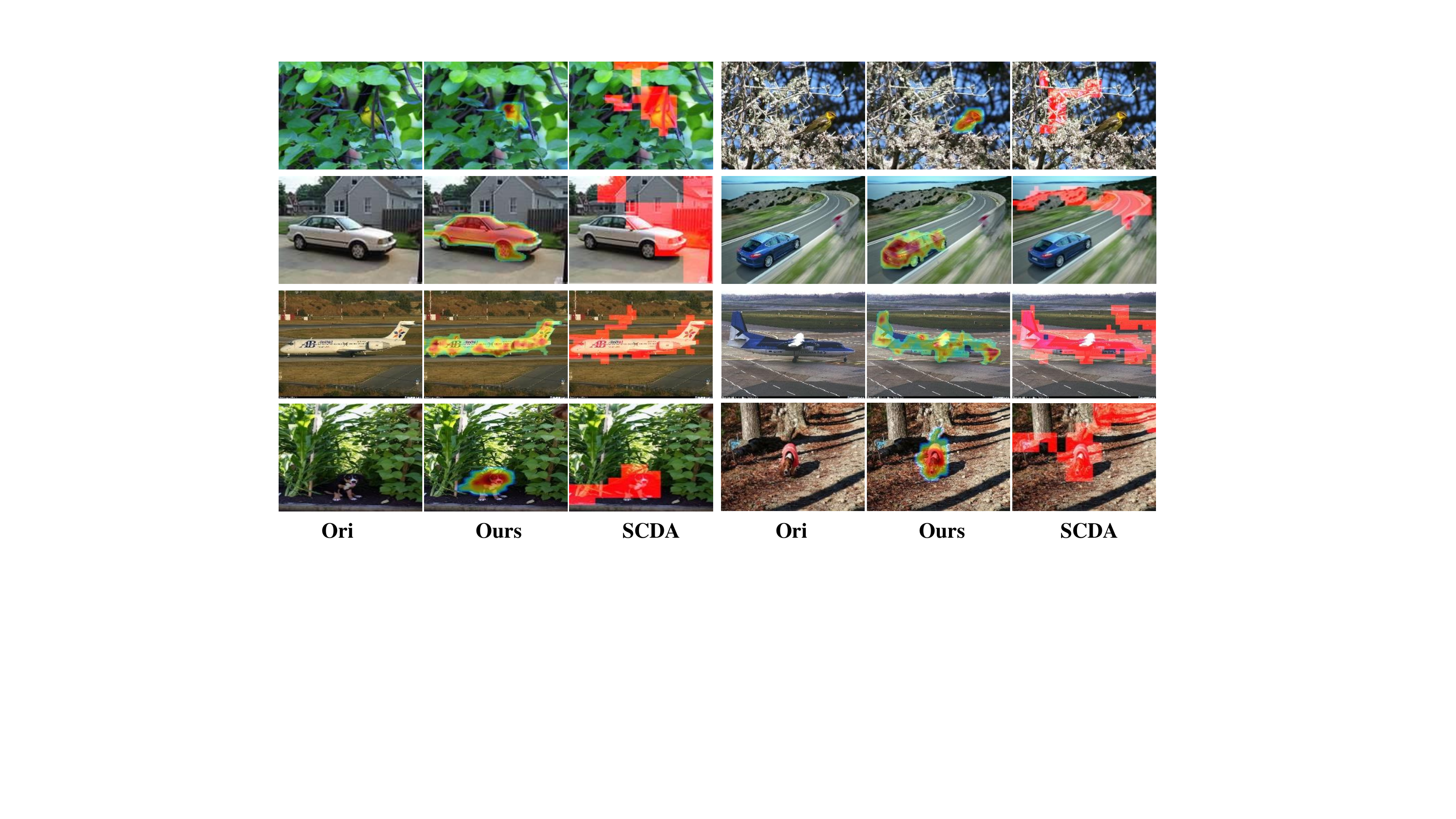}
\caption{Comparison with SCDA~\cite{Wei2017Selective} for generating localization maps in the same setting. Our method is more robust and can generate more accurate localization maps than SCDA~\cite{Wei2017Selective} (Best viewed in color.)
.}
\label{fig:scda}
\end{figure}

To verify the localization ability by introducing pattern mining techniques, we give the quantitative and the qualitative comparison with SCDA~\cite{Wei2017Selective}, which also reuses the pre-trained CNN model but employs a simple ``mean-threshod'' strategy to localize the objects. We detail the impact of our contributions on six datasets: CUB-200-2011, Standford Dogs, FGVC-aircrafts, Cars-196, VOC 2007 and ImageNet Subsets. The performances are reported in Table~\ref{tab:scda}. 
We can consistently observe that our pattern mining strategy achieves a significant improvement compared with~\cite{Wei2017Selective}. The superiority of our approach mainly benefits from introducing the pattern mining strategy to mine possible object regions. Specially, our approach outperforms~\cite{Wei2017Selective} by a large gain of 15.6\% and 22.3\% on two challenging datasets, VOC 2007 and ImageNet subsets, respectively. The results demonstrate that our method is more robust to complex images, which further validates the necessity of using pattern mining techniques. Figure~\ref{fig:scda} visualizes the localized regions of two methods. It can be seen that our approach obtains decent object localization results in many challenging scenarios.  Specifically, in the first rows, our method accurately locate the small objects. More interestingly, it discovers the precise objects even when they are very similar to the background or in a complex background. Meanwhile, our method can localize much finer contour of the objects compared with SCDA.  

\begin{table}[t]
\footnotesize
\renewcommand\arraystretch{1.2}
\setlength{\tabcolsep}{3pt}
\begin{tabular}{|c|c|c|c|c|c|c|}
\hline 
Methods&Aircrafts&Cars&Dogs&CUB&ImageNet&VOC2007\\
\hline\hline
\emph{Mean-threshold}\cite{Wei2017Selective}&94.9&91.0&78.9&76.8&37.8&29.5\\
\emph{Pattern mining }&\textbf{94.9}&\textbf{92.5}&\textbf{80.7}&\textbf{80.5}&\textbf{60.1}&\textbf{40.9}\\
\hline
\end{tabular}
\caption{Quantitative results with different strategy.}
\label{tab:scda}
\end{table}

\begin{figure}[ht]
\centering
\includegraphics[width=0.9\linewidth]{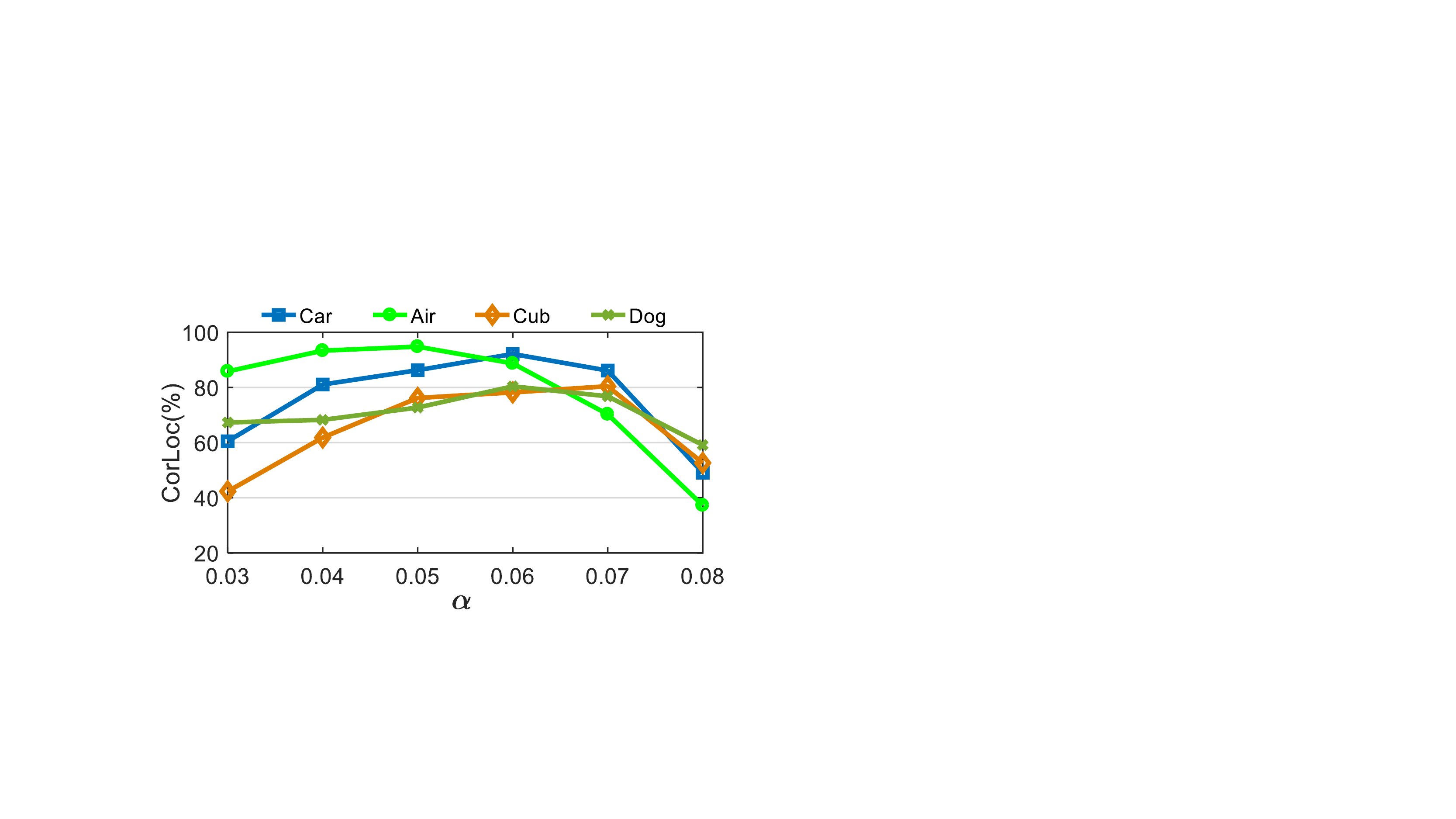}
\caption{ The localization performances based on different $\alpha$ on four fine-grained datasets.}
\label{fig:parameter}
\end{figure}

\begin{figure}[ht]
\centering
\includegraphics[width=1\linewidth]{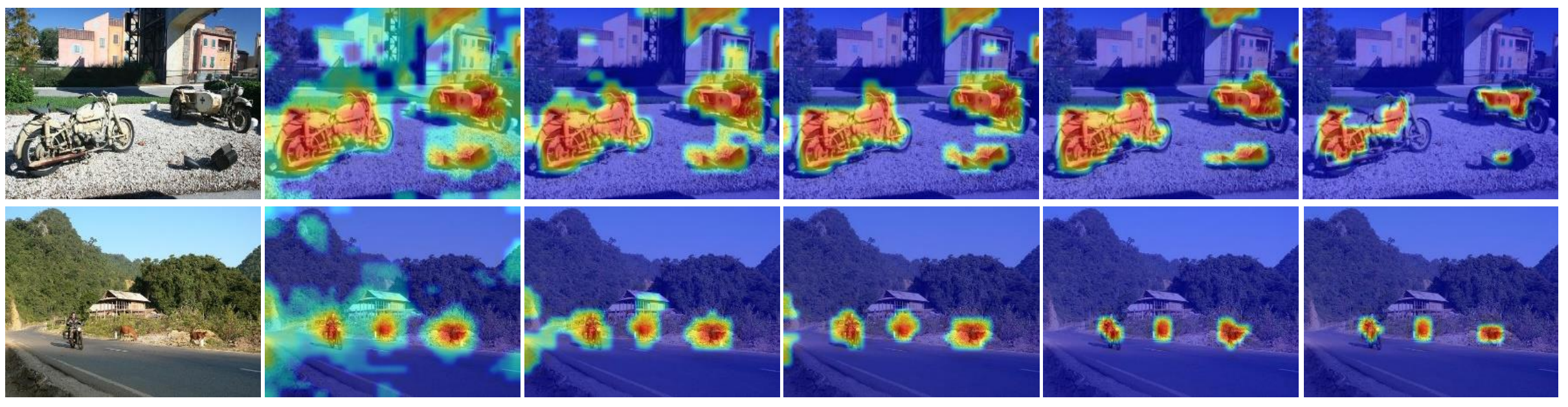}
\caption{Samples of the predicted support map based on the different $\alpha$. Note that $\alpha$ is gradually increasing from left to right.}
\label{fig:threshold}
\end{figure}

\subsubsection{Impact of parameter $\alpha$}
To investigate the effect of the parameter $\alpha$ on localization performances, we vary $\alpha$ from $0.03$ to $0.08$, finding that $\alpha=0.07$, $\alpha=0.06$, $\alpha=0.05$, $\alpha=0.06$ maximizes the accuracy on CUB-200-2011, Stanford Dogs, FGVC-Aircrafts and Stanford Cars-196 respectively, as
shown in Figure~\ref{fig:parameter}. The $\alpha$ is empirically set from 0.04 to 0.09 in our experiments.

In addition, we visualize some results with different threshold on challenging PASCAL VOC 2007 dataset, as shown in Figure~\ref{fig:threshold}. For example, for the first image, when $\alpha$ is set to a low value, the possible objects, including two motorbikes, the house and the light are all mined. But in most scenes, the house usually appears in the background, and the motorbikes often occur in the foreground. This causes the support value of motorbikes is higher than that of house. Thus, when $\alpha$ is set to a high value, the house is disappeared. The similar results can also been seen in the second image. We observe that a well-designed minimum support threshold $\alpha$ can improve the performance. Generally, a lower threshold can mine more possible objects but bring some background noise, while a higher threshold could reduce noise, but may only discover a small region of a target object.

\subsubsection{Strategy of convolutional feature combination} In order to apply pattern mining techniques successfully, we need carefully choose the patterns to contain all of the relevant information without information
loss. Meanwhile, we should also avoid the pattern explosion problem. Thus, we incorporate OLM feature produced from \emph{Relu-5} and \emph{Pool-5} in the VGG-16 model.
To verify the impact of features extracted from different convolutional layers, we perform the experiment with the following settings. \emph{Pool-5} refers to features maps extracted only with the \emph{Pool-5} layer, and \emph{ReLU-5} represents that features maps extracted only with the \emph{ReLU-5} layer. The localization results using different convolutional layers are reported in Table~\ref{tab:tab2}. Experimental results show that we can achieve significant improvement by combining feature maps extracted from \emph{Pool-5} and \emph{ReLU-5} layers on all four fine-grained datasets. Additionally, we also attempt to combine feature maps from different layers, \eg, \emph{Conv-4} and \emph{Pool-4}. However, the \emph{Conv-4}\&\emph{Pool-4} yields lower accuracy, and the cost of pattern mining method dramatically increases due to the rapid growth of items.

\begin{table}[t]
\footnotesize
\renewcommand\arraystretch{1.2}
\setlength{\tabcolsep}{7pt}
\begin{tabular}{|c|c|c|c|c|}
\hline 
Methods&CUB-200-2011&Dogs&Aircrafts&Cars-196\\
\hline\hline
\emph{Pool-5}& 76.64&79.58&93.40&87.30\\
\emph{ReLU-5}&65.54&67.33&92.24&84.32\\
\textbf{\emph{Pool-5}\&\emph{ReLU-5}}&\textbf{80.45}&\textbf{80.70}&\textbf{94.94}&\textbf{92.51}\\
\hline
\end{tabular}
\caption{Localization accuracy (\%) on four fine-grained datasets with different convolutional layers.}
\label{tab:tab2}
\end{table}

\subsection{Computational Complexity}
We randomly select 400 images from the CUB-200-2011 dataset as examples to report the computational complexity. We perform the experiments on a computer with Intel Xeon E5-2683 v3, 128G main memory, and a TITAN Xp GPU. Our proposed OLM approach consists of two major steps: (1) feature extraction and (2) pattern mining-based object localization including transaction creation, pattern mining, and support map generation. The execution time for feature extraction is about 0.03 second/image on GPU and 0.74 second/image on CPU, respectively. The second step only takes about 0.21 second/image on both GPU and CPU. Thus, the execute time is in total about 0.24 second/image on GPU and 0.95 second/image on CPU, respectively. This shows the efficiency of our proposed method in the practical scenario.

\section{Saliency Detection Experiments}

Our proposed method can also be used to detect salient regions. So we evaluate our proposed OLM on saliency detection task in this section.

\subsection{Datasets}
We evaluate OLM on seven datasets \ie~ESSSD~\cite{Yan2013HierarchicalESSSD}, PASCAL-S~\cite{Yin2014ThePASCALS}, THUR15K~\cite{ChengGroupSaliencyTHUR}, SOD~\cite{Jiang2013SalientDRFI}, SOC~\cite{fan2018SOC}, DAVIS~\cite{perazzi2016benchmark} and FBMS~\cite{brox2010object} for saliency detection.

\begin{itemize}
\item Image-based saliency datasets: We evaluate the proposed approach on five image-based saliency datasets, including ECSSD~\cite{Yan2013HierarchicalESSSD}, the PASCAL-S dataset~\cite{Yin2014ThePASCALS}, SOD \cite{Jiang2013SalientDRFI}, THUR15K~\cite{ChengGroupSaliencyTHUR} and the new SOC dataset~\cite{fan2018SOC}. ECSSD~\cite{Yan2013HierarchicalESSSD} contains 1000 images with multiple objects of different sizes and complex background. 
The PASCAL-S dataset~\cite{Yin2014ThePASCALS} ascends from the validation set of PASCAL VOC 2010~\cite{Everingham10} with 20 object
categories and complex scenes. THUR15K~\cite{ChengGroupSaliencyTHUR} contains 6,233 images about butterfly, coffee mug, dog jump, giraffe, and plane. SOD \cite{Jiang2013SalientDRFI}
 is a collection of salient object boundaries based on the Berkeley segmentation data set. This data set contains many images with multiple objects making it challenging. THUR15K contains 15K images with diverse and heterogeneous images from various internet sources. The SOC~\cite{fan2018SOC} dataset is the most challenging saliency detection dataset, which consists of saliency and non-saliency objects caused by motion, blurry, cluttered environments and occlusions, and thus can adequately reflect the complexity of images in the real world. 
 \item Video-based saliency datasets: In addition, we also evaluate our method on two video-based saliency datasets: DAVIS~\cite{perazzi2016benchmark} and FBMS~\cite{brox2010object}.
 \end{itemize}
 
\subsection{Evaluation Metric}
  We normalize the support map to [0-255] to generate the saliency map. To evaluate the proposed method, we utilize four metrics for quantitative performance evaluations following~\cite{wei2019f3net,fu2019deepside}, including F-measure, mean absolute error (MAE), S-measure~\cite{fan2017structure} and a recently proposed Enhanced-alignment metric, namely E-measure~\cite{Fan2018Enhanced}. We binarize the saliency maps with a threshold sliding from 0 to 255. They are briefly introduced as follows:
  \begin{itemize} 
  \item F-measure is a harmonic mean of average precision and average recall, formulated as:
   \begin{equation}
   F_{\beta }=(1+\beta^{2})\frac{Precision\times Recall}{\beta^{2}Precision+Recall},
   \end{equation}
  where $\beta^{2}$ is set to 0.3 as suggested in~\cite{wei2019f3net}.

   \item MAE refers to the average pixel-wise error between the saliency map and
   ground truth, which can be defined as:
  \begin{equation}
      MAE=\frac{1}{H}\sum_{h=1}^{H}\left | S(h)-GT(h)\right |,
  \end{equation}{}
  
  where H denotes the number of all pixels in the image. S and G denote the estimated saliency map and ground truth binary mask, respectively.
   \item S-measure ($S_{\theta }$) focuses on evaluating the structural information of saliency maps, which is closer to the human visual system than F-measure. S-measure could be computed as 
   \begin{equation}
       S_{\theta }=\theta * S_{0}+(1-\theta )*S_{\gamma, }
   \end{equation}
   
   where $S_{0}$ is the object-aware structural similarity and $S_{\gamma }$ is region-aware structural similarity. We set $\theta = 0.5$ as suggested in ~\cite{wei2019f3net,fan2017structure}.
   
    \item E-measure is proposed in \cite{Fan2018Enhanced} which consists of a single term to account for both pixel and image level properties.
\end{itemize}

\subsection{Image-based Saliency Detection Results}
We compare our proposed approach with 9 recent competitive unsupervised saliency models including FT~\cite{achanta2009frequency}, MC~\cite{Jiang2013SaliencyMC}, GMR~\cite{yang2013saliency} ,GC~\cite{cheng2013efficient}, PCA~\cite{margolin2013makes}, HS~\cite{yan2013hierarchical}, UFO~\cite{jiang2013salient}, BL~\cite{Tong2015} and BSCA~\cite{qin2015saliency}.
    
	For a fair comparison, we report the F-measure and MAE values on ECSSD, PASCAL-S, SOD and THUR15K dataset. The results are summarized in Table \ref{tab:saliency}. We can see that our method is competitive with other unsupervised methods on three popular datasets ( \eg, ECSSD, PASCAL-S and SOD) and consistently achieves the best MAE value on ECSSD and PASCAL-S datasets. Moreover, for the challenging datasets ( \eg , THUR15K dataset ), where the images are composed with complex backgrounds and diversified objects/targets, our proposed method ranks first in terms of both F-measure and MAE.
	
\begin{figure*}[ht]
\setlength{\abovecaptionskip}{0.15cm}
\begin{center}
\includegraphics[width=0.9\linewidth]{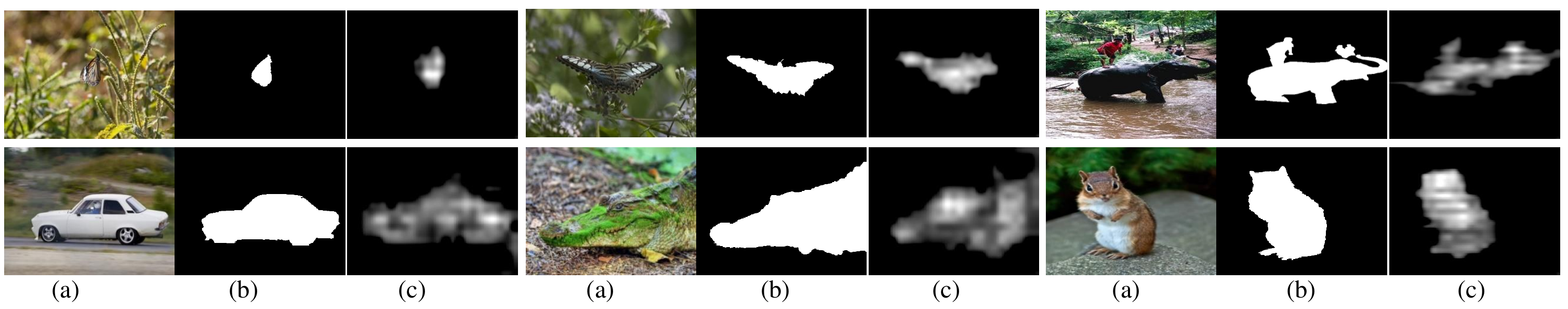}
   \caption{Examples of the saliency maps. (a) Input images. (b) Ground truth (c)Saliency maps achieved by our method.}
\label{fig:saliency}
\end{center}
\end{figure*}

	From the results, we can observe that our proposed method achieves better performance on more challenging datasets than relatively simple datasets. The reason might be that our method is mainly proposed for object localization task and thus prefers to exploit high-level semantic representation with low spatial resolution. This strategy leads to pay more attention to possible salient objects but loss some detailed low-level boundary information, which may reduce the accuracy of generated saliency map, especially on simple datasets. Some results of our method are shown in Figure~\ref{fig:saliency}, showing that we can achieve satisfactory saliency maps without any annotations.
	
  \begin{table*}[t]
   \scriptsize
    \centering
    \renewcommand\arraystretch{1.2}
    \setlength{\tabcolsep}{7.5pt}
    \begin{tabular}{|c|c|c|c|c|c|c|c|c|c|c|c|} 
    \hline 
   
    \multirow{2}{*}{Dataset}&\multirow{2}{*}{Metric}&\multicolumn{10}{c|}{Method}\\
    \cline{3-12} 
    &&FT~\cite{achanta2009frequency}&MC~\cite{Jiang2013SaliencyMC}&GMR~\cite{yang2013saliency}&GC~\cite{cheng2013efficient}&PCA~\cite{margolin2013makes}&HS~\cite{yan2013hierarchical}&UFO\cite{jiang2013salient}&BL~\cite{Tong2015}&BSCA~\cite{qin2015saliency}&OURS\\
    \hline
    \multirow{2}{*}{ECSSD}&F-measure~$\uparrow$&.449&.611&.693&.624&.580&.636&.644&.684&\textbf{.705}&.670\\
    &MAE~$\downarrow$&.285&.204&.237&.238&.290&.268&.205&.217&.217&\textbf{.201}\\
    \hline
    \multirow{2}{*}{PASCAL-S}&F-measure~$\uparrow$&.483&.574&.588&\textbf{.638}&.530&.528&.550&.574&.597&.625\\
    &MAE~$\downarrow$&.329&.272&.246&.265&.209&.223&.233&.249&.225&\textbf{.206}\\
   \hline
   \multirow{2}{*}{SOD}&F-measure~$\uparrow$&.441&.533&.570&.555&.537&.521&.548&.579&.534&\textbf{.597}\\
    &MAE~$\downarrow$&.283&.244&\textbf{.244}&.252&.274&.273&-&.267&.251&.254\\
    \hline
    \multirow{2}{*}{THUR15K}&F-measure~$\uparrow$&.386&.609&.597&.533&.544&.585&.579&.532&.536&\textbf{.634}\\
    &MAE~$\downarrow$&.177&.184&.142&.196&.198&.218&.165&.219&.183&\textbf{.132}\\
    \hline 
    \end{tabular}
    \caption{Comparison of maximum F-measure of average precision recall curve (F larger is better) 
    and MAE scores (M smaller is better)
    for different methods including ours on four benchmark datasets. The 1st score in each row is marked in bold.}
    \label{tab:saliency}
    \end{table*}

   Besides, the new SOC dataset~\cite{fan2018SOC} is the most challenging saliency detection benchmark, which includes images with salient and non-salient objects from daily object categories to reflect the real-world scenes in detail, so we conduct extensive evaluation on SOC dataset.
	Table \ref{tab:soc} reports a detailed comparison with 10 latest unsupervised methods published in recent five years. The quantitative saliency detection performance is evaluated by three evaluation metrics (\ie, S-measure, E-measure and MAE). We can see that the proposed method performs favorably against the recent state-of-the-art methods in terms of S-meaure and E-measure. Our method is only slightly lower than BFS~\cite{wang2015saliency1} in terms of MAE (.205 vs .195). These quantitative comparisons evidently verify the superiority of our proposed approach. More importantly, our method is simple which does not need any training process or prior information. It only takes less than 1 second per image to localize the object, which is discussed in Section IV-F ``Computational complexity''. The competitive results on SOC dataset demonstrates that incorporating pattern mining strategy with activations of pre-trained models is beneficial to object region discovery.
	
	\begin{table*}[t]
   \scriptsize
    \centering
    \renewcommand\arraystretch{1.2}
    \setlength{\tabcolsep}{5pt}
    \begin{tabular}{|c|c|c|c|c|c|c|c|c|c|c|c|c|}
    \hline 
    \multirow{2}{*}{Dataset}&\multirow{2}{*}{Metric}&\multicolumn{5}{c|}{2015-2016}&\multicolumn{5}{c|}{2017-2018}&\\
    \cline{3-13} 
    &&BL~\cite{Tong2015}&BSCA~\cite{qin2015saliency}&MAP~\cite{sun2015saliency}&BFS~\cite{wang2015saliency1}&DSP~\cite{chen2016discriminative}&WLRR~\cite{tang2016salient}&MIL~\cite{huang2017salient}&SMD~\cite{peng2016salient}&MDC~\cite{huang2017300}&WMR~\cite{zhu2018saliency}&OURS\\
    \hline
    \multirow{3}{*}{SOC-test}&S-measure~$\uparrow$&.623&.657&.644&.696&.644&.614&.671&.622&.675&.640&\textbf{.711}\\
    
    &E-mesure~$\uparrow$&.751&.755&.722&.753&.754&.759&.750&.748&.744&.733&\textbf{.770}\\
    
    &MAE~$\downarrow$&.296&.259&.256&\textbf{.195}&.248&.312&.236&.246&.219&.269&.205\\
    \hline 
    \end{tabular}
    \caption{Quantitative evaluations on the new SOC dataset. The 1st score in each row is marked in bold.} 
    \label{tab:soc}
    \end{table*} 

\subsection{Video-based Saliency Detection Results}	
  Since video saliency detection has attracted more attentions in recent years, we also evaluate the performance of our method on DAVIS and FBMS benchmark as suggested in~\cite{fan2019shifting},~\cite{wang2017saliency}, \cite{wang2017video}. As shown in Table~\ref{tab:saliency_video}, the proposed method is compared with 7 latest unsupervised video-based saliency algorithms including  MSTM~\cite{tu2016real}, STBP~\cite{xi2016salient}, SSA~\cite{li2017benchmark},
  GFVM~\cite{wang2015consistent}, SAGM~\cite{wang2015saliency}, RWRV~\cite{kim2015spatiotemporal} and three unsupervised image-based saliency algorithms including MB+M~\cite{zhang2015minimum}, RBD~\cite{Zhu2014SaliencyRBD} and MC~\cite{Jiang2013Saliency}. We adopt MAE, F-measure score, and S-measure score as our evaluation metrics. As shown in Table~\ref{tab:saliency_video}, the proposed method performs comparably or better than other 
  image-based unsupervised methods. Moreover, compared with all unsupervised methods including six video-based models, the proposed method considerably achieves the highest maxF and lowest MAE on the DAVIS dataset. Our method performs slightly worse than some unsupervised video-based models with respect to the S-measure metric. This may be caused by the fact that these models adopt temporal and motion information from videos while our method only utilizes the single frame.
  
\begin{table}[ht]
   \scriptsize
    \centering
    \renewcommand\arraystretch{1.1}
    \setlength{\tabcolsep}{8pt}
    \begin{tabular}{|c|c|c|c|c|c|c|} 
    \hline 
    \multirow{2}{*}{Method}&  
    \multicolumn{3}{c|}{FBMS}&\multicolumn{3}{c|}{DAVIS}\\
    \cline{2-7}  
    &F~$\uparrow$&M~$\downarrow$&S~$\uparrow$&F~$\uparrow$&M~$\downarrow$&S~$\uparrow$\\
    \hline 
    \multicolumn{7}{|c|}{Video-based unsupervised}\\
    \hline
    MSTM~\cite{tu2016real} &.500&.177&.613&.429&.125&.583\\
    STBP~\cite{xi2016salient} &.595&.152&.627&.544&.096&.677\\
    SSA~\cite{li2017benchmark} &.597&-&.634&.697&-&.738\\
    GFVM~\cite{wang2015consistent} &.571&.160&.651&.569&.103&.687\\
    SAGM~\cite{wang2015saliency} &.545&.161&.632&.515&.103&.676\\
    RWRV~\cite{kim2015spatiotemporal} &.336&.242&.521&.345&.199&.556\\
    \hline 
    \multicolumn{7}{|c|}{Image-based unsupervised}\\
    \hline
    MB+M~\cite{zhang2015minimum}&.487&.206&.609&.470&.177&.597\\
    RBD~\cite{Zhu2014SaliencyRBD} &.488&-&.591&.481&-&.620\\
    MC~\cite{Jiang2013Saliency} &.467&-&.567&.488&-&.590\\
    Ours  &.604&.340&.532&.512&.078&.672\\
    \hline 
    \end{tabular}
    \caption{Comparison of quantitative results using maximum F-measure, MAE, and S-measure. } 
    \label{tab:saliency_video}
    \end{table}

\vspace{-5 mm}
\section{Fine-Grained Classification}

Our proposed framework can also be easily applied to fine-grained classification task. Since the subtle visual differences among similar subcategories only locate at the discriminative parts, part localization is a key issue for fine-grained image classification. We observe that some relevant patterns generally correspond to discriminative part regions (\eg the head of bird) and the part regions are highlighted in the support map. Inspired by this observation, we can divide the regions into several groups of spatial locations. An intuitive idea is to perform the clustering algorithm on those mined patterns.

Specifically, we first produce the clustering data, which are three-dimensional data including the coordinates of each spatial location $(x,y)$ and its corresponding support map value $S(x,y)$. Then we take them as input of the k-means algorithm to cluster these connected regions into $K$ clusters. 
Surprisingly, the local regions represented by the patterns belonging to one cluster can be regarded as a discriminative part for a fine-grained image. Therefore, we obtain $K$ part locations $\textbf{C} = \{\textbf{c}_{1},\textbf{c}_{2},\ldots,\textbf{c}_{K}\}$ in the original image, where $\textbf{c}_{i}=(x_{i},y_{i})$ denotes the coordinates of the $i^{th}$ part.

After getting the part locations, $K$ parts are generated by cropping $K$ squares from $\textbf{I}$, with each element of $\textbf{C}$ as the square center. However, if the side length of the part square is simply fixed, some cropped parts may only include a small part but be disturbed by large background noises. In addition, a fixed-size part may lead to serious overlap with other parts. Therefore, in order to tackle the issues and generate more representative and distinctive parts, we consider a simple and effective geometric constrains to determine the side length $l$ of a part as follows:
\begin{equation}
l = \lambda \times min\{w_{o},h_{o}\}
\label{eqn:part},
\end{equation}
where $w_{o}$ and $h_{o}$ are width and height of the bounding box generated from the support map, respectively, and $\lambda$ is a scale factor. Finally, we can define the $i^{th}$ part mask as:
\begin{equation}
\textbf{M}_{i}(x,y) = \left\{
\begin{array}{ll}
1, & \text{if}~|x - x_{i}| \le \frac{l}{2}~\text{and}~|y - y_{i}| \le \frac{l}{2}\\ 
0, & \text{otherwise}\\
\end{array}.\label{eqn:part_mask}
\right.
\end{equation}
Thus, the $i^{th}$ part can be computed as follows:
\begin{equation}
\textbf{I}_{i}= \textbf{I} \odot \textbf{M}_{i},
\end{equation}
where $\odot$ denotes element-wise multiplication. 

\begin{figure}[t]
\setlength{\abovecaptionskip}{0.15cm}
\centering
\includegraphics[width=0.95\linewidth]{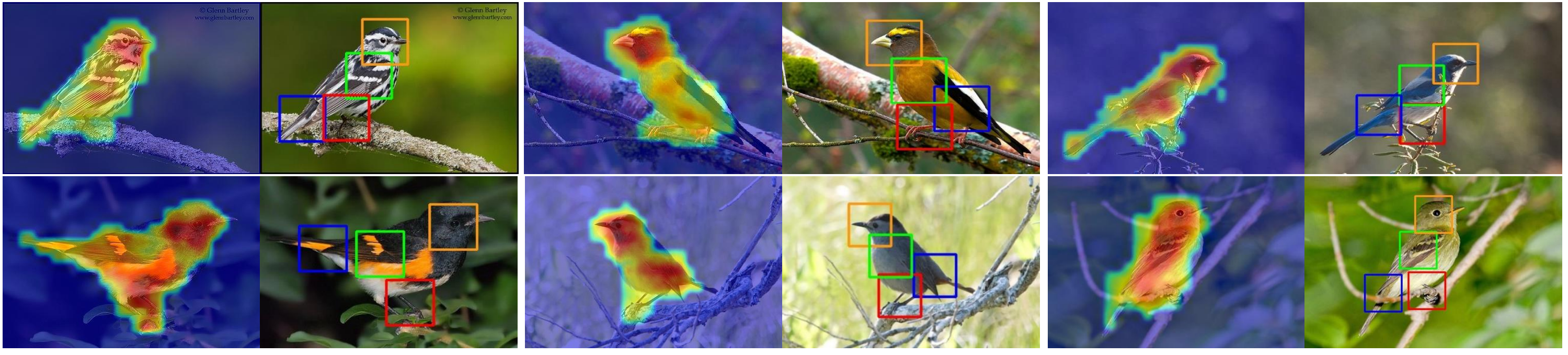}
\caption{Examples of the object locations and corresponding part localization results from CUB-200-2011. The yellow, green, blue and red bounding boxes dedicate the head, wing, tail and foot of the bird, respectively. (Best viewed in color)}
\label{fig:part_localization}
\end{figure}

Figure~\ref{fig:part_localization} shows some part localization results by our proposed approach.

Previous works~\cite{peng2018object,zheng2017learning} have proved the benefits of joint feature representation, thus we follow the same feature ensemble strategy. We build a multi-stream architecture, composed of \textit{Image stream}, \textit{Object stream} and \textit{Part stream}, with different level images as input. For the classification features of the original image, object and parts, the same CNN model is adopted but fine-tuned on different training data.
In our work, we follow~\cite{zheng2017learning} to obtain joint part-based representations for each image:
\begin{equation}
\{\textbf{P}_{I}, \textbf{P}_{O}, \textbf{P}_{1}, \textbf{P}_{2},\ldots, \textbf{P}_{K}\},
\end{equation}
where $\textbf{P}_{I}$, $\textbf{P}_{O}$ and $\textbf{P}_{i}$ denote the feature descriptors of the original image, the object image and the $i^{th}$ part, respectively. We extract the feature maps from the last convolutional layer of \textit{Image stream}, \textit{Object stream} and $K$ \textit{Part stream}, respectively. Then we perform the global average pooling and $\ell_{2}$-normalization to obtain the above feature descriptors. Finally, we concatenate them to train a classifier for the final classification.

For object/part localization, we use the VGG-16~\cite{Simonyan2014Very} model pre-trained on ImageNet with $448 \times 448$ resolution to extract feature maps from \emph{ReLU-5} and \emph{Pool-5} layers. The minimum support threshold $\alpha$ is set to 0.07. The number of parts $K$ is set to 4. The $\lambda$ in Equation~\ref{eqn:part} is empirically set as $\frac{1}{4}$ to produce more representative parts.

\begin{table}[htbp]
\footnotesize
\centering
\begin{tabular}{|c|c|} 
\hline 
Methods&Acc($\%$)\\
\hline  
\hline 
GoogLeNet-GAP~\cite{Zhou2016Learning} on full image& 63.0 \\
GoogLeNet-GAP~\cite{Zhou2016Learning} with object stream & 67.8\\
GoogLeNet-GAP~\cite{Zhou2016Learning} with BBox& 70.5\\
\hline
VGG16 on full image &69.7\\
VGG19 on full image &75.0\\
VGGNet-ACoL~\cite{XiaolinAdversarial2018} on full image & 71.9 \\
VGGNet-ADL~\cite{Choe_2019_CVPR} on full image & 65.3 \\
VGG16-Ours with object stream &72.3\\
VGG19-Ours with object stream& 76.9\\
VGG19-Ours with object stream+part stream& 82.5 \\
\hline
ResNet50 on full image&82.9 \\
ResNet50-GAP~\cite{Zhou2016Learning} on full image&80.6\\
ResNet50-ADL~\cite{Choe_2019_CVPR} on full image&80.3\\
ResNet50-Ours with object stream&83.6 \\
ResNet50-Ours with object stream+part stream &86.2 \\
\hline 
\end{tabular}
\caption{Classification results on CUB-200-2011.``Object'', and ``Parts'' represent whether the method uses the predict object annotations, part annotations.}
\label{tab:classfication}
\end{table}

Table~\ref{tab:classfication} summarizes the classification results on the CUB-200-2011 test set. To make fair comparison, we adopt VGG-16, VGG-19 and ResNet50 as baselines. It is noted that ACoL and ADL are the current state-of-the-art techniques for weakly supervised methods. In contrary, we do not use any annotations to localize object and part regions. In Table~\ref{tab:classfication}, we can observe that, using predicted objects annotations, the performance is improved by 2.6\% (69.7\% vs 72.3\% ) on VGG-16, 1.9\% (75.0\% vs 76.9\%) on VGG-19 and 0.7\% (82.9\% vs 83.6\%) on ResNet-50. Furthermore, benefited from the localized discriminative parts by our proposed approach, OLM (VGG-19 object+part) and OLM (ResNet-50 object+part) surpass two strong baseline models (VGG-19 on full image and ResNet-50 on full image) by 7.5\% and 3.6\% significant improvement, respectively.
Compared with~\cite{Choe_2019_CVPR,XiaolinAdversarial2018,Zhou2016Learning}, OLM outperforms them by a noticeable margin.
We achieve better classification results than VGGNet-ACoL~\cite{XiaolinAdversarial2018} (82.5\% vs 71.90\%) and VGGNet-ADL~\cite{Choe_2019_CVPR} (82.5\% vs 65.3\%) respectively. OLM (ResNet-50) also achieves the best result among all of the ResNet-50 based methods. 

The reason is that our mined regions can locate the discriminative parts and thus boost the classification performance. Moreover, unlike other
techniques~\cite{Choe_2019_CVPR,XiaolinAdversarial2018,Zhou2016Learning}, our localization approach does not need training process, thus it is very fast to extract objects and parts regions, and can be easily integrated with various classification models.

\section{Conclusion}
In this paper, we propose a novel pattern mining-based method, called Object Location Mining (OLM), for object discovery and localization from a single unlabeled image. Our method exploits the advantage of data mining and feature representation of pre-trained CNN models. The proposed OLM can also be easily extended to the task of saliency detection and fine-grained classification. Experimental results show that OLM achieves competitive performance on a variety of benchmarks, demonstrating the effectiveness of coupling pattern mining with pre-trained model reuse. Our approach does not need any annotations yet still shows promoising localization ability, which provides a new perspective to solve the localization problem.

\ifCLASSOPTIONcaptionsoff
  \newpage
\fi





\bibliographystyle{IEEEtran}
\bibliography{IEEEabrv,Bibliography}
%

\vspace{-10 mm}
\begin{IEEEbiography}[{\includegraphics[width=1in,height=1.25in,clip,keepaspectratio]{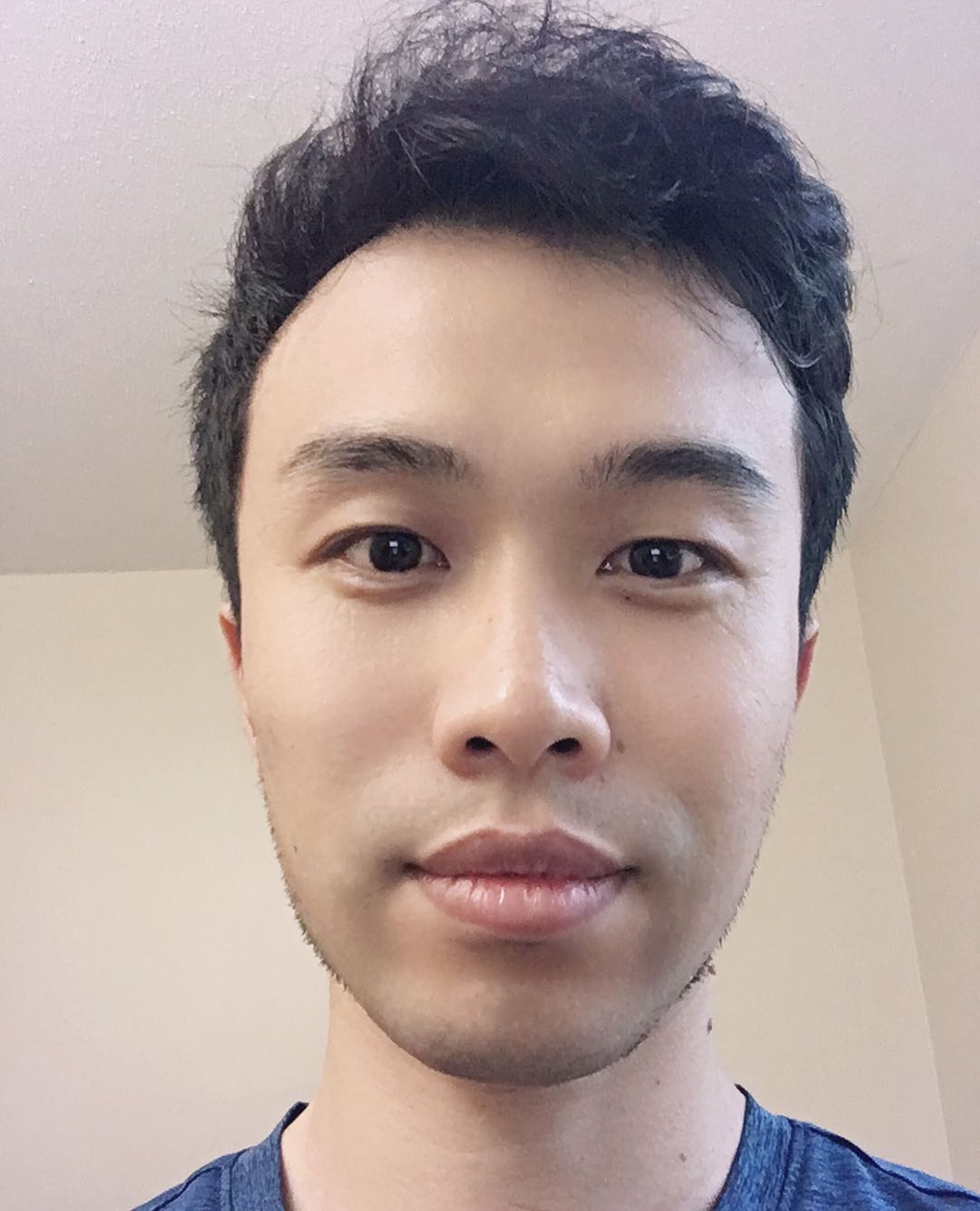}}]
{Runsheng Zhang} received his B.S. and M.S. degrees in the School of Information Science and Engineering at Shenyang Ligong University. He is currently a Ph.D. candidate in the School of Computer and Information Technology at Beijing Jiaotong University, supervised by Prof. Yaping Huang. His research interests include object discovery, data mining and computer vision.
\end{IEEEbiography}
\begin{IEEEbiography}[{\includegraphics[width=1in,height=1.25in,clip,keepaspectratio]{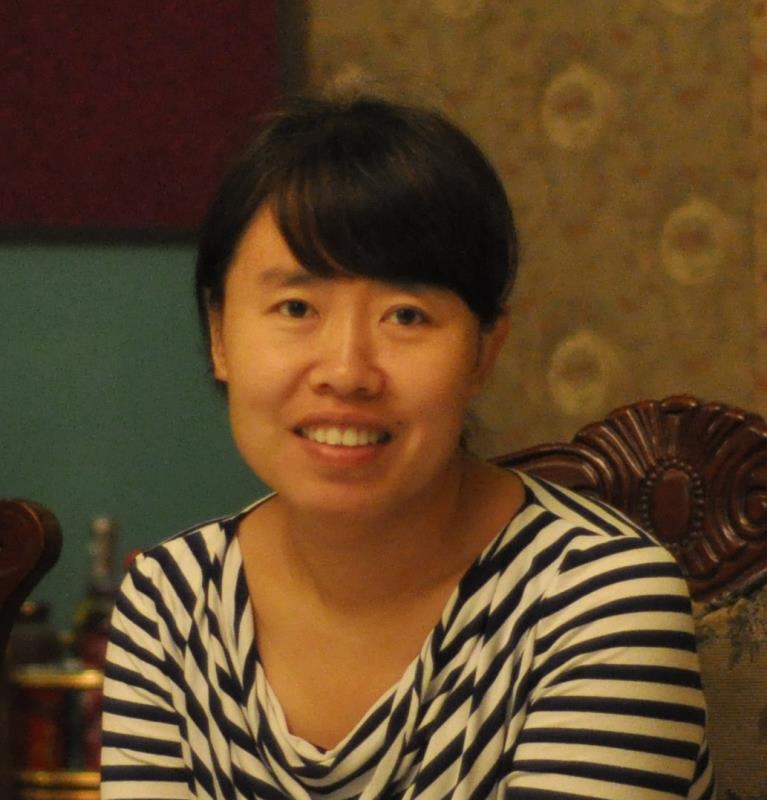}}]{Yaping Huang} received the B.S., M.S., and Ph.D. degrees from the School of Computer and Information Technology, Beijing Jiaotong University, China, in 1995, 1998, and 2004, respectively. She is currently a professor with the School of Computer and Information Technology, Beijing Jiaotong University. Her research interests include computer vision, machine learning, pattern recognition.
\end{IEEEbiography}
\begin{IEEEbiography}[{\includegraphics[width=1in,height=1.25in,clip,keepaspectratio]{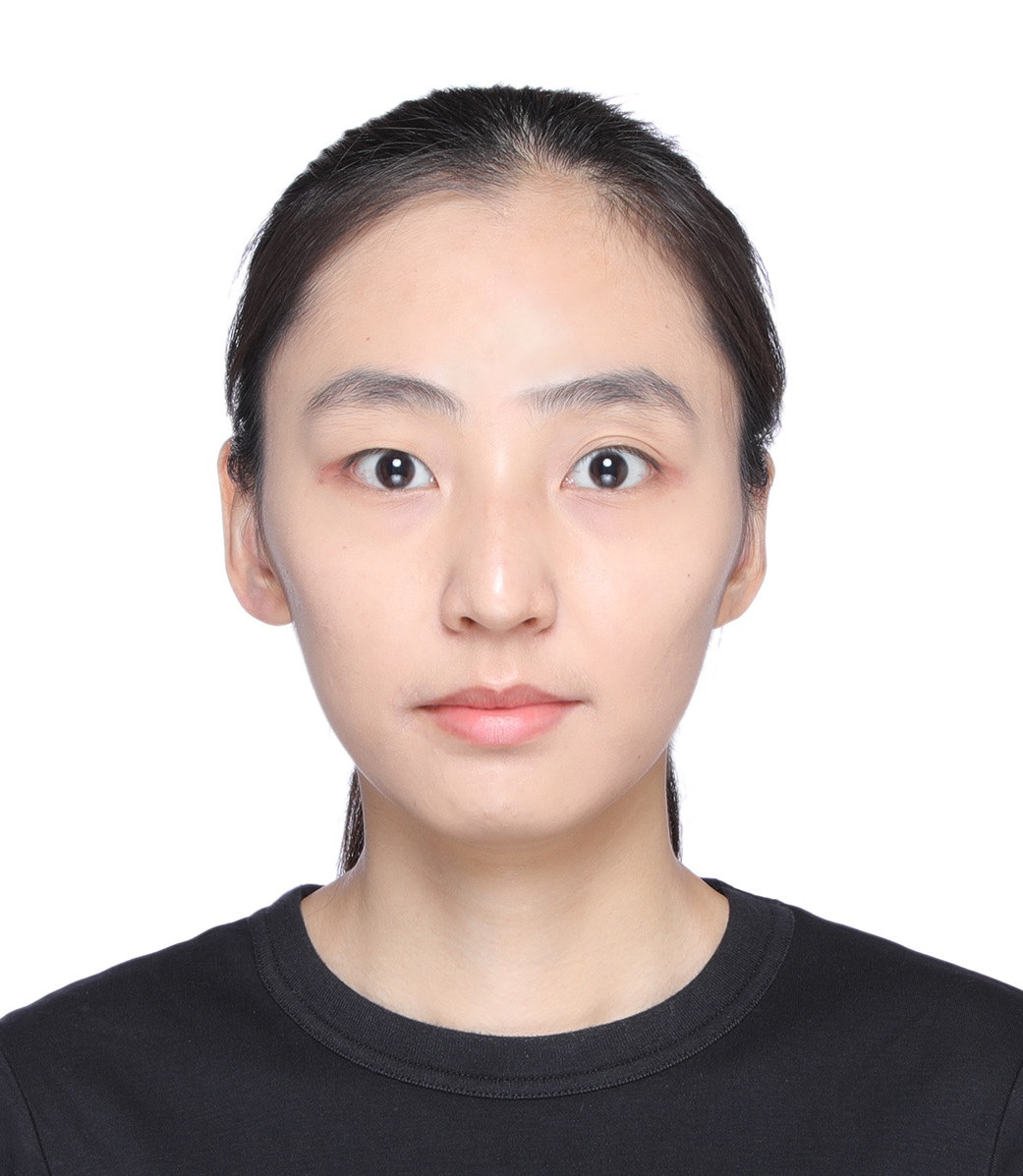}}]{Mengyang Pu} is a Ph.D. candidate in the School of Computer and Information Technology at Beijing Jiaotong University. Her current research interests include computer vision, weakly-supervised semantic segmentation and localization.
\end{IEEEbiography}

\begin{IEEEbiography}[{\includegraphics[width=1in,height=1.25in,clip,keepaspectratio]{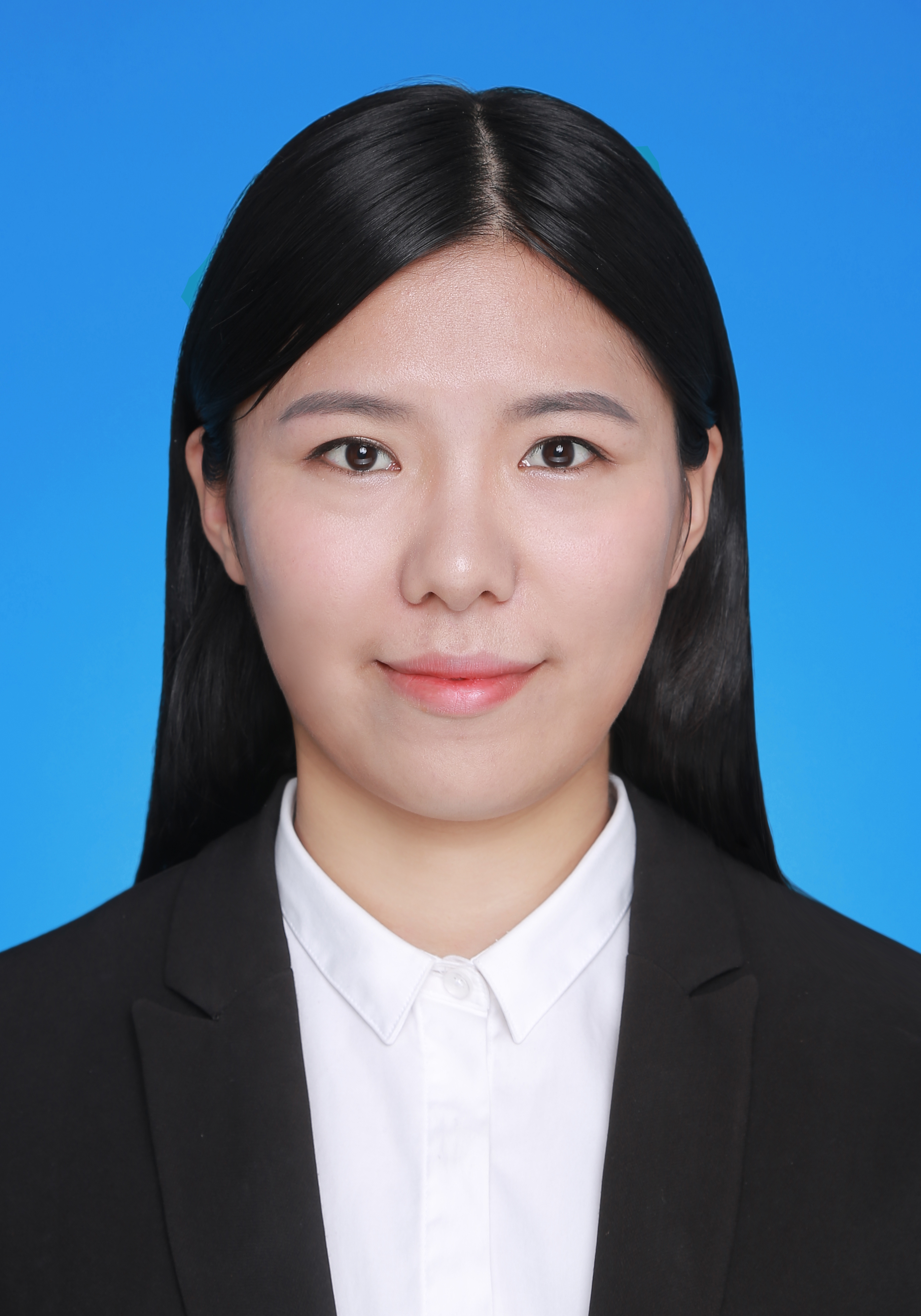}}]{Jian Zhang} received her B.S. and M.S. degrees in the School of Computer and Information Technology at Beijing Jiaotong University in 2017 and 2020. Her research interests are computer vision and pattern recognition.
\end{IEEEbiography}

\begin{IEEEbiography}[{\includegraphics[width=1in,height=1.25in,clip,keepaspectratio]{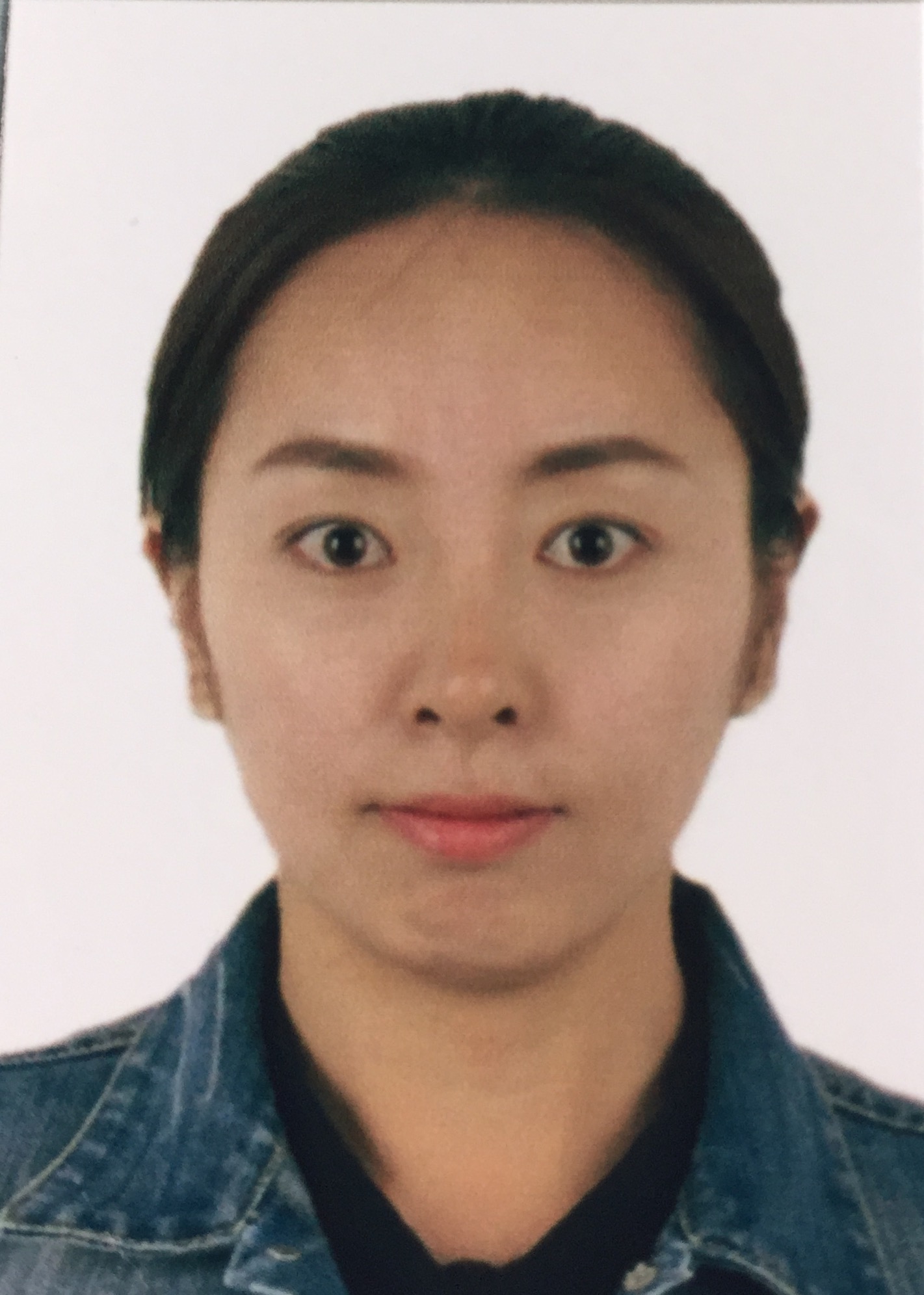}}]{Qingji Guan} received her B.S. and M.S. degrees in Computer Science and Technology from Northeast Normal University in 2012 and 2015. She is currently a Ph.D. candidate in the School of Computer and Information Technology at Beijing Jiaotong University, supervised by Prof. Yaping Huang. Her research interests include computer vision, medical imaging.
\end{IEEEbiography}

\begin{IEEEbiography}[{\includegraphics[width=1in,height=1.25in,clip,keepaspectratio]{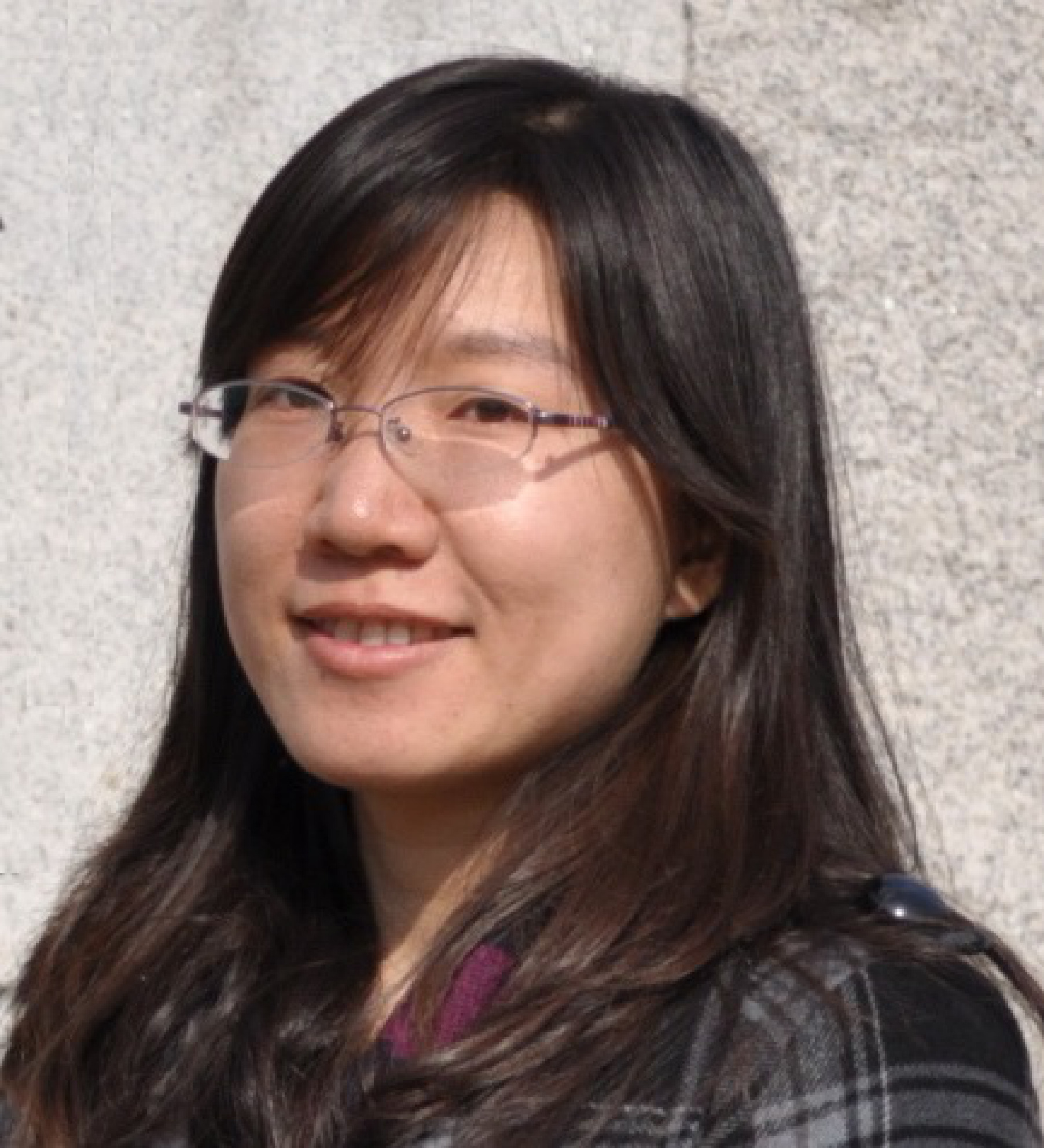}}]{Qi Zou}received her B.S., and Ph.D. degree from Beijing Jiaotong University, China in 2001 and 2006 respectively. She is currently a professor in the School of Computer and Information Technology at Beijing Jiaotong University. Her research interests include pattern recognition, computer vision and target tracking.
\end{IEEEbiography}

\begin{IEEEbiography}[{\includegraphics[width=1in,height=1.25in,clip,keepaspectratio]{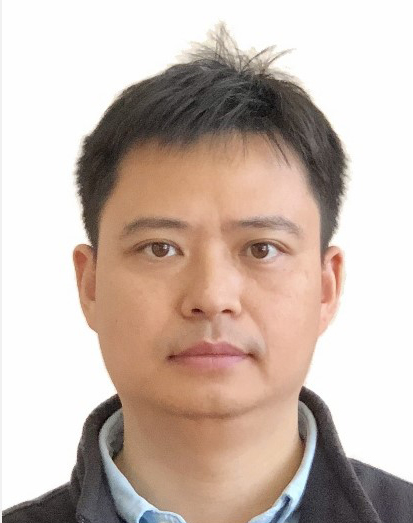}}]{Haibin Ling }received the B.S. degree in mathematics and the M.S. degree in computer science from Peking University, China, in  1997 and 2000, respectively, and the Ph.D. degree from the University of Maryland, College Park, in Computer Science in 2006. From 2000 to 2001, he was an assistant researcher at Microsoft Research Asia. From 2006 to 2007, he worked as a postdoctoral scientist at the University of California Los Angeles. After that, he joined Siemens Corporate Research as a research scientist. He then joined Temple University as an Assistant Professor in 2008  and was later promoted to Associate Professor. In fall 2019, he joined SUNY Stony Brook University as an Empire Innovation Professor. He is an Associate Editor of IEEE Trans. on PAMI, Pattern Recognition, and CVIU, and served as Area Chairs for CVPR2014, 2016 and 2019.
\end{IEEEbiography}





\vfill


\end{document}